\documentclass[sigconf,nonacm]{acmart}

\AtBeginDocument{%
  }

\setcopyright{acmlicensed}
\copyrightyear{2018}
\acmYear{2018}
\acmDOI{XXXXXXX.XXXXXXX}

\acmConference[Conference acronym 'XX]{Make sure to enter the correct
  conference title from your rights confirmation email}{June 03--05,
  2018}{Woodstock, NY}

\usepackage{amsmath}
\usepackage{booktabs}
\usepackage{array}
\usepackage{tabularx}
\usepackage{fvextra}
\usepackage{float}
\usepackage{longtable}
\usepackage{pdflscape}
\usepackage{colortbl}
\usepackage{wrapfig}
\usepackage{supertabular}
\usepackage{booktabs}
\usepackage{capt-of}
\usepackage{needspace}
\usepackage[utf8]{inputenc}
\makeatletter
\AddToHook{cmd/x@supertabular/after}
   {%
     \let\orig@startpbox@action\@startpbox@action
     \let\@startpbox@action\@startpbox
   }
\AddToHook{cmd/ST@restore/after}
   {%
     \let\@startpbox@action\orig@startpbox@action
   }
\makeatother

\usepackage[most]{tcolorbox}
\usepackage{adjustbox}
\usepackage{ragged2e}
\usepackage{fontawesome5}
\definecolor{blockgray}{gray}{0.95}
\definecolor{ftblue}{RGB}{235, 241, 246}
\definecolor{ftaccent}{RGB}{40, 75, 110}
 
\newtcolorbox{promptbox}[2][]{
    colback=ftblue,
    colframe=ftaccent,
    rounded corners,
    arc=2mm,
    boxrule=0.4pt,
    borderline west={2pt}{0pt}{blue!70!black},
    fonttitle=\bfseries,
    title={{#2}},
    before upper=\ttfamily,
    breakable,
    width=\linewidth,
    #1
}

\author{
    Snehalkumar `Neil' S. Gaikwad \textsuperscript{1,2*}\textsuperscript{\faEnvelope[regular]},
    Dhruv Dixit\textsuperscript{2*}, 
    Bryen Alvarez\textsuperscript{1*}, 
    Pratik Patekar\textsuperscript{2*},\\
    Anika Sharma\textsuperscript{1*}, 
    Akshara Kolipaka\textsuperscript{1*}, 
    Paritosh Pandey\textsuperscript{1*}
}
\affiliation{\vspace{0.5em}
  \institution{\textsuperscript{*}Society-Centered AI Lab}
  \institution{\textsuperscript{1}Department of Computer Science}
  \institution{\textsuperscript{2}School of Data and Information Sciences}
  \city{University of North Carolina, Chapel Hill}
  \country{USA}
}
\email{gaikwad@cs.unc.edu}

\renewcommand{\shortauthors}{Gaikwad et al.}
\acmISBN{978-1-4503-XXXX-X/2018/06}

\begin{document}

\title[AI-Voyager]{AI-Voyager: A Public-Interest System for Human-AI Civic Reasoning to Contest and Shape the Rules That Govern AI}

\begin{abstract}
Governments are writing rules for artificial intelligence that decide what counts as an automated decision or a harm, often before the people they affect have a chance to question them. We introduce AI-Voyager, a public-interest system for human-AI civic reasoning, in which people and AI work together to understand, contest and shape the rules that govern AI. A resident can find their city's rules on AI in hiring or upload their country's AI strategy, read it as an affected worker would, open the passage behind any answer, and draft a revision. Its architecture turns 3,653 governance documents from city, state, federal and national governments into passages that keep their page and edition, and retrieves evidence through bounded agentic search that verifies each statement. On 360 new questions that require finding the relevant documents, it reached 83.0\% macro recall, against 80.5\% for the strongest baseline, 73.7\% without iterative search and 99.1\% on a benchmark with fixed documents. In a study with 24 participants new to two AI bills, most proposed changes asked how provisions would apply. Participants chose lenses for the kind of explanation they wanted, and citations signaled grounding but rarely led them to check the source. By binding AI assistance to the exact text of public documents, this work introduces a grammar of operations and a design space that inform how such systems are designed and evaluated, and how people take part in governing AI.
\end{abstract}



\begin{teaserfigure}
\centering
\includegraphics[width=0.95\linewidth]{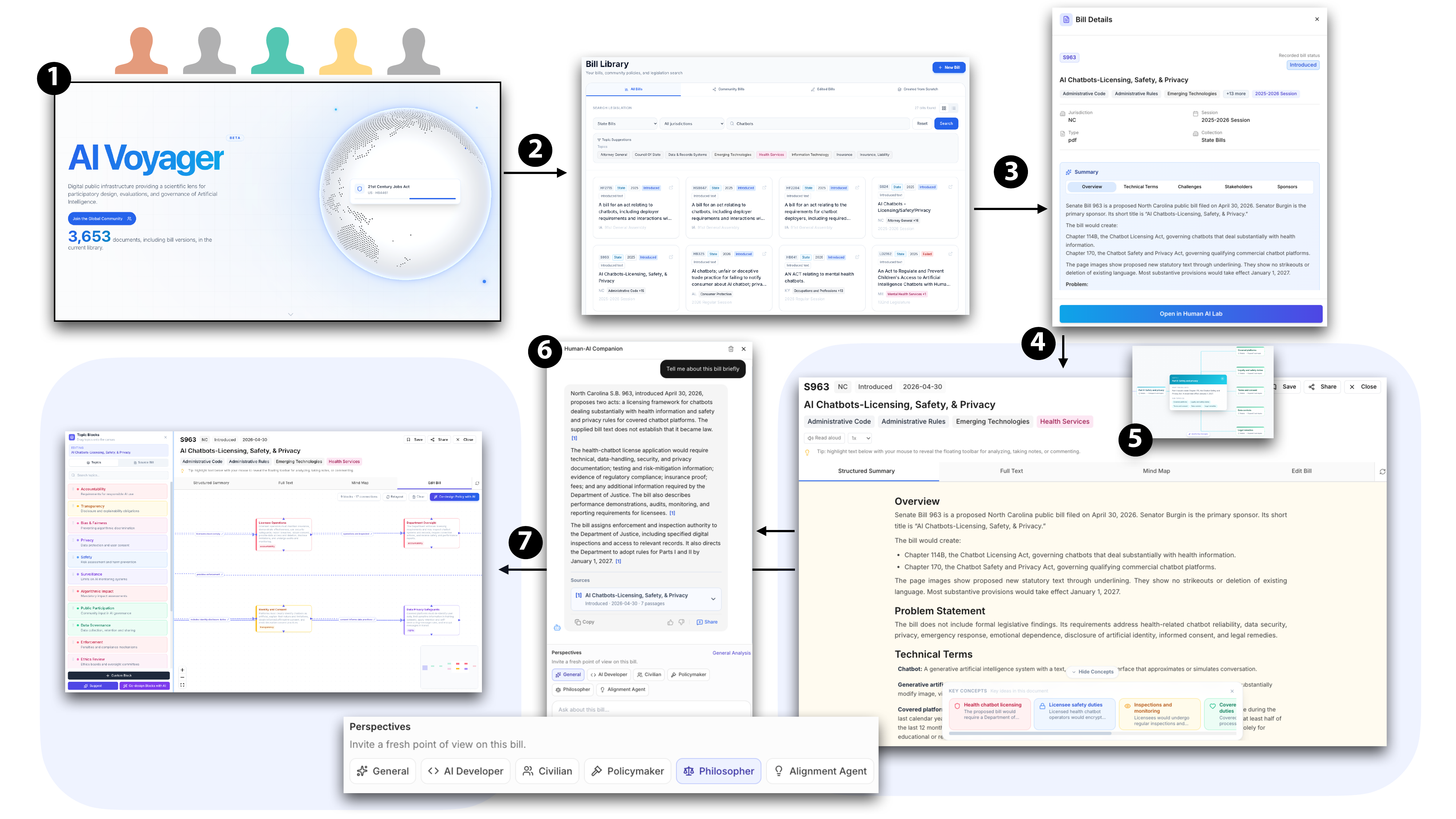}
\caption{\textbf{AI-Voyager takes a person from a search across legislation to a cited answer and an edited bill.} \textbf{1}, The landing page opens a library of 3,653 documents, including bill versions. \textbf{2}, The Bill Library filters bills by collection, jurisdiction, topic, and search term, here ``Chatbots.'' \textbf{3}, Bill Details shows a selected bill, North Carolina S963, with its status, session, and a tabbed summary. \textbf{4}, The Human AI Lab presents the bill as a structured summary, full text, mind map, and editing view, with key concepts drawn from the document. \textbf{5}, The mind map lays out the bill visually. \textbf{6}, The Human-AI Companion answers questions with numbered citations to the bill's passages. A person can choose a perspective: General, AI Developer, Civilian, Policymaker, Philosopher, or Alignment Agent. \textbf{7}, The Edit Bill canvas arranges topic blocks, such as Accountability, Transparency, and Privacy, into a connected policy that a person can co-design with AI.} 
\Description{Three-panel architecture diagram. Panel a shows the interface and nine operations people perform. Panel b shows the five-step Voyager-RAG loop with verification. Panel c shows the five corpus collections, the ingestion pipeline, and corpus scale.}
\label{fig:architecture}
\end{teaserfigure}

\maketitle

\section{Introduction}

How might people anywhere in the world take part in deciding the rules that govern AI? AI systems increasingly shape the decisions public institutions make about people's lives~\cite{alon2024public, johnson2025legacy}, and a growing body of work documents the harms they cause~\cite{raji2020closing, costanza2022audits}. Responding to those harms means deciding who is responsible, what protections apply, and how affected people can question decisions~\cite{cobbe2023understanding}, choices that encode positions on fairness, accountability, and whose interests count~\cite{johnson2022bureaucratic, selbst2019fairness, abdu2024algorithmic}. AI legislation makes these choices early. It decides what counts as an automated decision, what counts as harm, and whose experience is relevant, often before anyone knows who will be affected. Alignment research has begun to bring public input into the principles that guide model behavior~\cite{bai2022constitutional, huang2024collective}, and participatory AI research lets stakeholders state preferences about algorithmic policy~\cite{delgado2023participatory, birhane2022power, suresh2024participation, lee2019webuildai}. We treat societal alignment as work in other direction: articulating, contesting, and revising the rules that AI systems and the institutions deploying them should follow~\cite{bogiatzis2024beyond, stone2026legitimate, shen2024towards}. Members of the public can contribute to this work by finding where a law is underspecified, provided they can reach its exact text.

Reaching the exact text at scale is hard, and an AI system that helps people reach it becomes part of what they see. Governance text is fragmented across jurisdictions, institutions, and editions~\cite{nannini2023explainability}, mixes enacted requirements with proposals and plans~\cite{schiff2020s}, and arrives as scanned and born-digital PDFs, slides, web pages, and spreadsheets. The meaning of a provision depends on its edition and on definitions and exceptions stored in other sections. Retrieval-augmented generation (RAG) provides source passages to a language model~\cite{lewis2020retrieval}, yet a relevant passage can still fail to support the statement it is cited for~\cite{gao2023enabling}, and graph-based indexes~\cite{edge2024local} must keep qualifications intact~\cite{campi2026mder} as documents change. The AI system raises questions of its own. Participation already faces power imbalances~\cite{sloane2022participation, groves2023going, buckner2026challenges}, an accurate and well-cited interpretation can still be mistaken for what the policy establishes~\cite{ehsan2021expanding}, and stakeholder perspectives generated by a model~\cite{sorensen2024roadmap} cannot speak for the communities they name~\cite{agnew2024illusion, karusala2024understanding, yurrita2025identifying}. We call this activity human-AI civic reasoning: people and AI working together to understand, contest, and shape the rules that govern AI. A system for it must keep every claim traceable to one passage in one edition across thousands of changing documents, and it must be studied as an actor in the reasoning it supports.

In this paper, we introduce \textbf{AI-Voyager}, a public-interest system for human-AI civic reasoning (Figure~\ref{fig:architecture}). It runs on a corpus we collected from scratch: 2,318 policy records in 3,653 document versions and 278,021 pages from city, state, federal, and national governments, spanning 21 policy domains. Its interface gives people a grammar of operations that they combine into their own reasoning. They find a policy or bring their own jurisdiction's document, read it, ask about it with an AI Companion bound to that document version, reframe it through five analytical lenses, verify a claim by opening its cited passage, and revise or publish a proposal. 

We evaluate the system in two ways. Existing legal retrieval benchmarks fix the documents in advance, while public questions start further back: the relevant bills must be found, the right edition chosen, and evidence assembled across documents. On a new benchmark of 360 source-reviewed questions that include these steps, Voyager-RAG reaches 83.0\% macro recall at fifteen passages, against 80.5\% for the strongest baseline and 99.1\% on LegalBench-RAG-mini~\cite{pipitone2024legalbench}, where documents are fixed. Removing iterative retrieval or claim verification lowers macro recall to 73.7\% and 76.5\%. In a within-subjects study, 24 participants (including 6 pilot studies) with little familiarity with two AI bills explored them with and without the lenses. Asked what they would change, they mostly asked how provisions would apply: what a bill covers, how its licensing and data retention would work, and who would enforce it. Participants chose lenses for the kind of explanation they wanted as well as the viewpoint, and citations signaled grounding but rarely led them to check the source. Together, these results outline a design space for human-AI civic reasoning (Section~\ref{sec:discussion}). We make following contributions:

\begin{itemize}
    \item \textbf{AI-Voyager, a public-interest system for human-AI civic reasoning}: a hosted production system whose grammar of operations lets people compose their own reasoning over the exact text of policy, running on a corpus collected from scratch across city, state, federal, and national governments and 21 policy domains.
    \item \textbf{A provenance-preserving architecture} that reconstructs 278,021 pages in reading order into 812,537 passages bound to their version, page, and character span, with bounded agentic retrieval in which statement verification drives evidence acquisition.
    \item \textbf{Technical and human evaluations}: a benchmark of 360 source-reviewed governance questions that covers document discovery and cross-document evidence, with human validation and BM25, dense, and hybrid baselines, and a study showing that non-experts ask how AI bills would apply in practice while the system's lenses, suggested questions, and citations shape what they explore and trust.
    \item \textbf{A design space for human-AI civic reasoning}, with eight dimensions across technical, human-AI, and societal layers and implications for how such systems are aligned, measured, and evaluated.
\end{itemize}

\section{Related Work}
\label{sec:related-work}

Taking part in AI governance requires finding the policies that apply, reading them from more than one position, checking interpretations against the text, and proposing revisions. We review work on each of these steps, from governance and participation to the two technical areas AI-Voyager builds on.

\subsection{AI Governance and Public Access to Policy}

AI governance connects technical systems to the institutions and communities affected by their use~\cite{alon2024public}. AI ethics initiatives vary in authorship, motivation, and intended effect~\cite{schiff2020s}, accountability depends on actors across supply chains and deployment contexts~\cite{cobbe2023understanding, lewicki2023out}, and algorithmic governance must account for existing institutional rules and for how accountability passes between institutions~\cite{johnson2022bureaucratic, selbst2019fairness, abdu2024algorithmic}. Scholars argue for collective democratic authority over AI~\cite{bogiatzis2024beyond} and for collaboration between HCI and policymaking~\cite{yang2024future}, and COMPL-AI translates EU AI Act requirements into model evaluations for developers~\cite{guldimann2024compl}. Policy trackers make AI policy visible at scale. The OECD.AI Policy Observatory catalogues national strategies~\cite{oecdai}, trackers follow AI bills across U.S. states~\cite{iappTracker}, and OpenStates and LegiScan publish bill texts and status~\cite{openstates, legiscan}. Interactive tools support public budgets~\cite{kim2016budgetmap}, climate decisions~\cite{cao2024llm}, public discourse~\cite{wang2025policypulse}, and legislative proposals~\cite{yun2025legigpt}. AI-Voyager builds on these resources by taking users from a catalogue entry to the exact document version that a citation refers to.

\subsection{Public Participation in AI Policy and Lawmaking}

Participatory AI research centers stakeholder agency~\cite{delgado2023participatory, birhane2022power, lupetti2026participatory}, cautions against superficial or extractive participation~\cite{sloane2022participation, buckner2026challenges}, calls for methods suited to public-sector institutions~\cite{saxena2025emerging}, and identifies the information needs of people contesting public-sector decisions~\cite{karusala2024understanding, yurrita2025identifying}. Preference-elicitation systems model stakeholder preferences for algorithmic policy~\cite{lee2019webuildai, noothigattu2018voting}. Crowdlaw projects bring the public into lawmaking: Regulation Room helped non-experts comment on proposed U.S. federal rules~\cite{r2013regulation}, Finland crowdsourced input for its off-road traffic law~\cite{aitamurto2016crowdsourced}, and Polis, used in Taiwan's vTaiwan process, maps opinion groups to surface consensus~\cite{small2021polis}. Civic deliberation tools ground discussion in shared materials~\cite{reynante2021framework, palombo2025framing, kabbara2025ai, fan2020digital, peng2025glitter}, and authoring tools support revising the rules for AI models and chatbots~\cite{feng2026policypad, lam2024ai, petridis2024constitutionmaker}. AI-Voyager brings these strands together for governance documents across jurisdictions, from reading a specific bill version to publishing a proposed revision.

\begin{table*}[t]
\centering
\setlength{\tabcolsep}{4pt}
\renewcommand{\arraystretch}{1.5}
\caption{The operations of AI-Voyager's grammar of civic reasoning.}
\label{tab:grammar}
\begin{tabular}{@{}p{1.4cm}lp{7cm}p{5cm}@{}}
\toprule
\textbf{Operation} & \textbf{Design Goal} & \textbf{What a person does} & \textbf{Interface} \\
\midrule
Find    & DG2 & Locate a policy by place, keyword, topic, or question & World Map, Bill Library, AI Chat \\
Bring   & DG2 & Add a document from their own jurisdiction & Upload \\
Read    & DG1 & Read the text and its reading aids & Full text, summary, mind map \\
Ask     & DG4 & Question one version or the whole collection & AI Companion, AI Chat \\
Reframe & DG3 & View the same source from another position & Lenses \\
Verify  & DG4 & Open a cited passage in its source & Expandable citations \\
Compare & DG1 & Set versions or jurisdictions side by side & AI Chat, version records \\
Revise  & DG5 & Draft changes against the original wording & Edit Bill \\
Publish & DG5 & Share observations and proposals & Launchpad \\
\bottomrule
\end{tabular}
\end{table*}

\subsection{LLM Perspectives, Personas, and Pluralistic Alignment}

Human-AI interaction research shows that people need to understand an AI system's capabilities before working with it~\cite{cai2019hello, amershi2019guidelines, hughes2026design}, that the configuration of human and algorithmic decision-makers shapes perceived fairness~\cite{yurrita2025towards}, that explanation depends on its human context~\cite{ehsan2021expanding}, and that conversational interfaces can expose uncertainty to support democratic inquiry~\cite{delacroix2025designing}. Persona prompting is best understood as role play~\cite{shanahan2023roleplay}. Demographic backstories can reproduce some survey distributions~\cite{argyle2023out}, while default model opinions diverge from those of many groups~\cite{santurkar2023opinions}, models standing in for identity groups misportray and flatten them~\cite{wang2025large}, and personas do not improve performance on objective tasks~\cite{zheng2024personas}. Multi-agent debate uses several viewpoints to reach one answer~\cite{du2024debate}. Constitutional AI uses explicit principles~\cite{bai2022constitutional}, Collective Constitutional AI draws them from public input~\cite{huang2024collective}, pluralistic models can present a spectrum of responses or be steered to one perspective~\cite{sorensen2024roadmap}, and alignment also runs from AI to people~\cite{shen2024towards}. We apply several perspectives to one fixed source text that a person then reads and acts on. AI-Voyager's lenses are analytical viewpoints bound to the same document version and citations, and they make no claim to represent stakeholder communities~\cite{agnew2024illusion}.

\subsection{Retrieval-Augmented Generation and Verifiable Legal Evidence}

RAG conditions model outputs on external passages~\cite{lewis2020retrieval}. In law, general-purpose models hallucinate about court cases~\cite{dahl2024fictions}, and commercial legal research tools still produce misgrounded answers~\cite{magesh2025hallucination}. Graph and hierarchical methods organize evidence into derived structures~\cite{edge2024local, zhao2026hierarchical} whose strengths depend on the task~\cite{han2025rag}, and adaptive and agentic methods let a model decide when and what to retrieve~\cite{jiang2023flare, asai2024selfrag, yan2024crag, singh2025agentic}. Governance text adds two requirements: qualifications can be lost when text is reduced to graph triples~\cite{campi2026mder}, and bills are amended and reintroduced, so derived structures must track every version. Benchmarks with fixed documents measure legal reasoning~\cite{guha2023legalbench, fei2024lawbench}, legal retrieval~\cite{pipitone2024legalbench}, and citation quality~\cite{gao2023enabling}, and generative search engines often cite sources that do not support their statements~\cite{liu2023verifiability}. People over-rely on AI recommendations~\cite{bucinca2021trust}, and explanations reduce overreliance mainly when they make verification cheaper~\cite{vasconcelos2023overreliance}. Documentation and oversight research stresses provenance and access to supporting information~\cite{gebru2021datasheets, mitchell2019model, crisan2022interactive, boyd2021datasheets, sterz2024quest, raji2020closing, costanza2022audits}. AI-Voyager keeps evidence at the level of source passages with provenance, bounds its agentic retrieval, and is evaluated where documents must first be found.

\section{Designing AI-Voyager}
\label{sec:design}

A system for human-AI civic reasoning has to decide how people find the right document and edition, what the AI says and what it shows, how to keep a bill's text apart from interpretations and proposals, whether perspectives share the same evidence, and how a reader can propose and share a change. We derived five design goals from the prior work in Section~\ref{sec:related-work} and from iterative development of the system.

\subsection{Design Goals}
\label{sec:design-goals}

\textbf{DG1. Make the governance record legible.}
People need to know which institution issued a policy, whether it is enacted, proposed, or a plan, and which edition they are reading~\cite{schiff2020s, nannini2023explainability, johnson2022bureaucratic}. AI-Voyager keeps policy records distinct from document versions, shows jurisdiction, status, and date with every document, and gives each edition its own citation number.

\textbf{DG2. Support discovery without legislative expertise.}
Members of the public rarely know bill numbers or statutory terms, and civic interfaces lower this barrier by organizing information around issues people recognize~\cite{kim2016budgetmap, palombo2025framing}. AI-Voyager offers geographic browsing, keyword search, topic suggestions, metadata filters, and plain-language questions across the collection.

\textbf{DG3. Offer several perspectives on one shared source.}
Several perspectives can widen what a reader considers~\cite{sorensen2024roadmap}, and model-generated perspectives can also be mistaken for the voices of the groups they name~\cite{agnew2024illusion, wang2025large}. AI-Voyager presents its lenses as analytical viewpoints and binds every lens to the same document version and citations.

\textbf{DG4. Keep facts verifiable and distinct from interpretation.}
Language models often misstate the law~\cite{dahl2024fictions, magesh2025hallucination}, a cited passage can fail to support its claim~\cite{gao2023enabling, liu2023verifiability}, and an accurate interpretation can still be mistaken for what a policy establishes~\cite{ehsan2021expanding}. In AI-Voyager, only indexed source passages count as evidence, and summaries, mind maps, and discussion messages serve as reading aids. Every factual statement carries identifiers that open its supporting passage, statements without support are removed, and responses keep factual explanation separate from interpretation.

\textbf{DG5. Support revision while leaving judgments to the user.}
Participation is more meaningful when people can develop, examine, and revise the positions they contribute~\cite{delgado2023participatory, lee2019webuildai, feng2026policypad}. AI-Voyager lets users draft amendments against the original wording or write new bills, keep versions, and publish proposals for discussion. The AI labels the changes it suggests as proposals, and users decide what to keep and share.

The goals leave out legal advice, enforcement tracking, and speaking for stakeholder communities. Table~\ref{tab:dg-map} maps each goal to the features and components that implement it and to the evaluation that tests it.

\begin{table*}[t]
\centering
\renewcommand{\arraystretch}{1.5}
\setlength{\tabcolsep}{3pt}
\caption{How each design goal is implemented and evaluated.}
\label{tab:dg-map}
\begin{tabularx}{\linewidth}{@{}
  >{\raggedright\arraybackslash}X
  >{\raggedright\arraybackslash}X
  >{\raggedright\arraybackslash}X
  >{\raggedright\arraybackslash}X
@{}}
\toprule
\textbf{Design goal} &
\textbf{Interface} &
\textbf{Technical component} &
\textbf{Evaluation} \\
\midrule
DG1. Legible governance record &
Version header, status labels, per-edition citation numbers &
Record and version model, canonical catalogue
(\S\ref{sec:representation}) &
Version-specific gold sets
(\S\ref{sec:benchmark}) \\
DG2. Discovery without expertise &
World Map, Bill Library, AI Chat, document upload &
Scope resolution, dense search, source-balanced retrieval
(\S\ref{sec:rag-system}) &
120 discovery questions
(\S\ref{sec:benchmark}) \\
DG3. Perspectives on one source &
Five lenses and General Analysis &
Lenses share one version scope and evidence pool
(\S\ref{sec:rag-system}) &
User study (\S\ref{sec:study}) \\
DG4. Verifiable facts, distinct from interpretation &
Expandable citations, evidence windows &
Passage provenance
(\S\ref{sec:representation}), statement verification
(\S\ref{sec:rag-system}) &
Benchmarks, ablations
(\S\ref{sec:evaluation}), user study
(\S\ref{sec:study}) \\
DG5. Revision with judgments left to the user &
Edit Bill, Launchpad &
Proposal labeling in the reviewer
(\S\ref{sec:rag-system}), versioned drafts &
Not evaluated directly; participants' proposed changes
(\S\ref{sec:study}) \\
\bottomrule
\end{tabularx}
\end{table*}

\subsection{A Grammar of Civic Reasoning}
\label{sec:grammar}

AI-Voyager gives people a small grammar of objects and operations that they combine into their own path through a law, much as the grammar of graphics composes charts from a few parts~\cite{wilkinson2011grammar}. The objects are policy records, document versions, passages with their provenance, citations, lenses, drafts, proposals, and uploaded documents, and nine operations act on them (Table~\ref{tab:grammar}). Because every operation works on versioned passages, any path a person takes stays traceable to the source. The system supplies the operations, and people decide which to use and in what order.

\subsection{The AI-Voyager Interface}

\begin{figure*}[t]
    \centering
    \includegraphics[width=\linewidth]{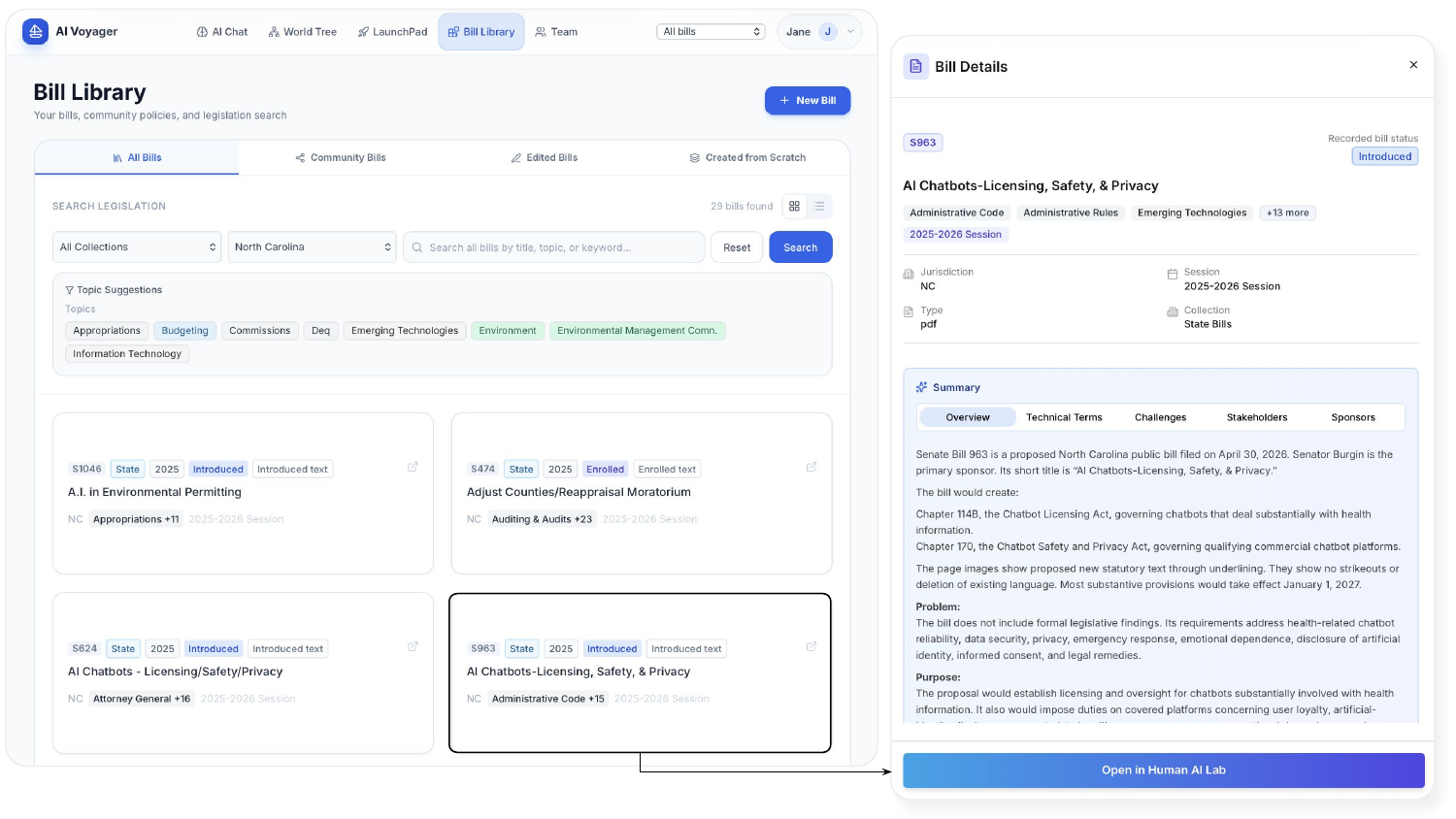}
    \caption{\textbf{AI-Voyager's Bill Library.} (a) Users search AI governance legislation across jurisdictions and filter by collection, jurisdiction, topic, and keyword. Each bill card shows the bill's identifier and legislative status. (b) Selecting a bill opens its details panel, shown here for North Carolina Senate Bill 963. The panel lists the bill's title, status, session, and collection, followed by a summary covering its overview, technical terms, challenges, stakeholders, and sponsors. From the panel, users can open the bill in the Human AI Laboratory to read and analyze it with the AI Companion.}
    \label{fig:bill-library}
\end{figure*}

We describe the interface through Maya, a volunteer at a community organization who has been asked to explain North Carolina Senate Bill 963, a proposed bill on AI chatbots, to residents.\footnote{Her reading and analysis steps mirror the task in our study (Section~\ref{sec:study}), which did not include Edit Bill or Launchpad.} To \emph{Find} it, she can browse a World Map of countries, states, and cities or search the Bill Library, which filters by collection and jurisdiction and shows each bill's status and indexed version (DG1, DG2; Figure~\ref{fig:bill-library}). A reader elsewhere could \emph{Bring} a governance document from their own jurisdiction by uploading it, so the system is not limited to the documents its designers collected.

Opening the bill shows the Human AI Laboratory (Figure~\ref{fig:architecture}), with a structured summary, the full text with highlighting, notes, and comments, a mind map, and the Edit Bill workspace. Maya \emph{Asks} the AI Companion what user consent the bill requires. The Companion answers about the selected version only and offers suggested questions on the bill's overview, technical requirements, impact, and enforcement. Its answers group citations by version, and a marker such as [1] expands into passage references such as [1.1] that open the evidence in its source location, so Maya can \emph{Verify} the consent duty on page 2 of the introduced edition (DG4). She then \emph{Reframes} the bill (DG3). The AI Developer lens examines implementation and failure modes, Civilian everyday effects and access, Policymaker legal mechanisms and institutional responsibilities, Philosopher rights, duties, consequences, and fairness, and Alignment Agent the tools and citizen-facing services the provisions could support. AI Chat answers questions across the collection, so she can \emph{Compare} how other jurisdictions treat the same issue.

Unsure what counts as explicit consent, Maya \emph{Revises} the bill in Edit Bill, which imports the source text as editable topic blocks, offers model assistance, compares the draft with the original, and saves versions. She \emph{Publishes} her clarification to Launchpad, where users share observations and drafts for comment and take part in guided policy challenges (DG5). Her path combines seven of the grammar's nine operations, in an order she chose. Edit Bill and Launchpad have been demonstrated at public exhibitions, where we collected no data, and our study does not cover them.

\section{Provenance-Preserving Representation and Retrieval}
\label{sec:architecture}

Three components keep AI-Voyager's factual claims traceable: a pipeline that reconstructs documents from their sources (Section~\ref{sec:corpus}), a representation that ties every passage to one location in one edition (Section~\ref{sec:representation}), and a retrieval system that gathers and checks evidence within that representation (Section~\ref{sec:rag-system}). The design rests on reading-order reconstruction with completeness checks, reversible normalization that keeps qualifications with their provisions, scope restrictions that bind every retrieval round, and statement verification that can trigger further search. Table~\ref{tab:corpus} gives the scale of each component. We follow one question throughout: a reader of North Carolina Senate Bill 963 asks what user consent the bill requires. The answer is clause (b)(3) of the proposed \S114B-4, which takes its meaning from the list of operational requirements it belongs to.

\subsection{Corpus Construction with ROAR}
\label{sec:corpus}

We collected the corpus from scratch in five collections: U.S. federal and state bills, city policies, national legislation from outside the United States, and national strategies (Table~\ref{tab:corpus}). Federal bills supply most of the pages, averaging 136 pages per version, while state bills carry the most revision history, with 2.3 versions per record. Binding AI legislation is concentrated in the United States, while most other countries have so far addressed AI through national strategies~\cite{stanford_ai_index_2025}.

\begin{table*}[t]
\centering
\small
\setlength{\tabcolsep}{5pt}
\renewcommand{\arraystretch}{1.1}
\caption{AI-Voyager corpus composition, document and passage statistics, and system configuration. Passage lengths count embedding-input tokens.}
\label{tab:corpus}
\begin{minipage}{0.96\textwidth}
\textbf{Corpus composition by collection}\par\vspace{0.15em}
\begin{tabular*}{\linewidth}{@{}l@{\extracolsep{\fill}}rrrr@{}}
\toprule
Collection & Records & Versions & Pages & Passages \\
\midrule
U.S. federal bills   & 1,404 & 1,799 & 244,194 & 635,863 \\
U.S. state bills     & 707   & 1,647 & 16,379  & 70,274 \\
National strategies  & 152   & 152   & 11,306  & 65,288 \\
National legislation & 29    & 29    & 5,721   & 39,392 \\
City policies        & 26    & 26    & 421     & 1,720 \\
\midrule
\textbf{Total} & \textbf{2,318} & \textbf{3,653} & \textbf{278,021} & \textbf{812,537} \\
\bottomrule
\end{tabular*}
\vspace{0.65em}

\noindent
\begin{minipage}[t]{0.485\linewidth}
\textbf{Document and passage statistics}\par\vspace{0.15em}
\begin{tabularx}{\linewidth}{@{}>{\raggedright\arraybackslash}Xr@{}}
\toprule
Metric & Value \\
\midrule
Declared physical pages & 278,021 \\
Available source pages & 277,436 \\
Median pages per document version & 9 \\
95th-percentile document length & 271 pages \\
Maximum document length & 5,432 pages \\
Versions exceeding 1,000 pages & 77 \\
\addlinespace[0.35em]
Mean passage length & 103.8 tokens \\
Median passage length & 58 tokens \\
95th-percentile passage length & 353 tokens \\
99th-percentile passage length & 506 tokens \\
Passages per document version (mean) & 222.4 \\
Total embedding-input tokens & 84,341,942 \\
\bottomrule
\end{tabularx}
\end{minipage}\hfill%
\begin{minipage}[t]{0.485\linewidth}
\textbf{Corpus scale and system configuration}\par\vspace{0.15em}
\begin{tabularx}{\linewidth}{@{}>{\raggedright\arraybackslash}Xr@{}}
\toprule
Quantity & Value \\
\midrule
Policy records / document versions & 2,318 / 3,653 \\
Physical pages reconstructed & 278,021 \\
Passages / embedding tokens & 812,537 / 84.3M \\
\addlinespace[0.35em]
\multicolumn{2}{@{}l}{\textit{Ingestion and representation}} \\
GPUs / concurrent document workers & 1 A100 / 16 \\
Provenance fields per passage & 6 \\
Embedding dimensions & 1,536 \\
\addlinespace[0.35em]
\multicolumn{2}{@{}l}{\textit{Retrieval}} \\
Retrieval rounds (maximum) & 4 \\
Candidates per round (scope + named) & 15 + 15 \\
Named documents balanced (maximum) & 4 \\
Evidence budget (tokens) & 10,000 \\
\bottomrule
\end{tabularx}
\end{minipage}
\end{minipage}
\end{table*}

\paragraph{Challenge.}
Serializing a PDF's embedded text can interleave columns, detach headings from their clauses, and flatten tables, so a definition can merge with a neighboring provision before retrieval begins. Scanned pages have no embedded text, and printed page labels rarely match the physical page numbers a citation needs.

\paragraph{ROAR transcription.}
Reading Order-Aware Reconstruction (ROAR) sends every document through one path. Each physical page is rendered with PDFium at 200 DPI, resized to a maximum edge of 1,540 pixels, and transcribed in visual reading order by LightOnOCR-2-1B~\cite{taghadouini2026lightonocr} with greedy decoding and up to 8,192 output tokens. The same procedure applies to born-digital pages, scanned pages, and pages with existing OCR layers, so no document depends on the quality of its embedded text. 

Postprocessing converts each transcription into heading, paragraph, list, and table blocks, keeps row and column spans for merged table cells that flat Markdown cannot express, and stores the raw output beside the normalized text. A page counts as complete only if its transcription is not truncated, empty, or repetitive, and near-white pages are recorded as blank. Physical page numbers are stored separately from printed labels. Native adapters handle HTML, DOCX, PPTX, spreadsheets, and plain text.

\paragraph{Ingestion.}
Ingestion of all 278,021 pages ran on a single NVIDIA A100 80\,GB GPU shared by 16 document workers through NVIDIA Multi-Process Service~\cite{nvidia_mps}. Every source file receives a SHA-256 hash, and a coordinator commits each completed page to a persistent checkpoint before acknowledging the worker. A resumed run reuses a page only when its source, model, backend, and extraction identities all match, so an interrupted ingestion never mixes outputs from different configurations. In the running example, ROAR keeps page 2 in reading order, so clause (b)(3) stays under the heading of the requirements it belongs to.

\paragraph{Corpus statistics.}
Table~\ref{tab:corpus} reports document and passage lengths. Document length is highly skewed: the median version has 9 pages, while 77 versions exceed 1,000 pages and the longest has 5,432. The 585-page difference between declared physical pages and available source pages comprises 578 pages containing only headers or footers and seven image-only pages without text.

\begin{figure*}[t]
    \centering \includegraphics[width=0.9\linewidth]{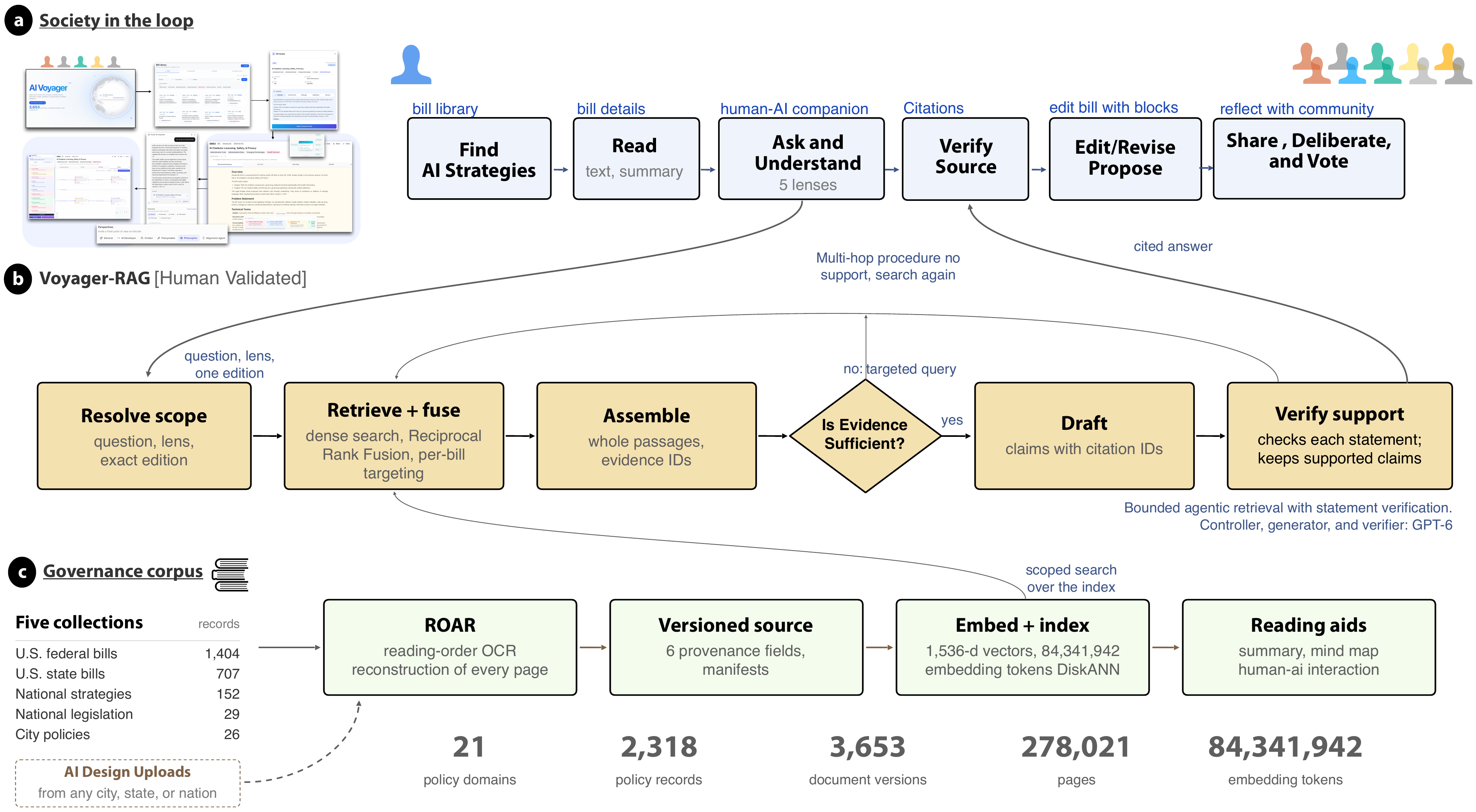}
    \caption{\textbf{AI-Voyager's architecture for human-AI civic reasoning.} \textbf{(a) Society in the loop:} Users find governance documents, read them, ask questions, verify cited evidence, revise provisions, and share proposals for public discussion. They can combine these steps freely and view legislation through different analytical lenses. \textbf{(b) Voyager-RAG:} For each question, the system identifies which documents are in scope, retrieves and fuses relevant passages, and assembles the evidence used to generate an answer. An agentic controller decides whether more retrieval is needed. A verifier checks each generated statement against its cited evidence and can trigger targeted retrieval if statement lacks support. \textbf{(c) Governance corpus:} The ROAR pipeline processes five collections of governance documents, along with user-uploaded policies, into versioned passages that keep their source provenance and are embedded for retrieval. Summaries, mind maps, and key concepts serve as additional reading aids.}    \Description{Schematic Representation of AI-Voyager}
        \label{AI-Voyager_schematic}
\end{figure*}

\subsection{Knowledge Representation for Governance Text}
\label{sec:representation}

\paragraph{Challenge.}
Fixed-length chunking can cut a definition from its conditions, and extracting a graph of entities and relations can drop the qualifications that decide coverage~\cite{campi2026mder}. Governance documents are also amended, so any derived structure must stay aligned with each new version~\cite{edge2024local, zhao2026hierarchical}.

\begin{figure*}[t]
    \centering \includegraphics[width=0.9\linewidth]{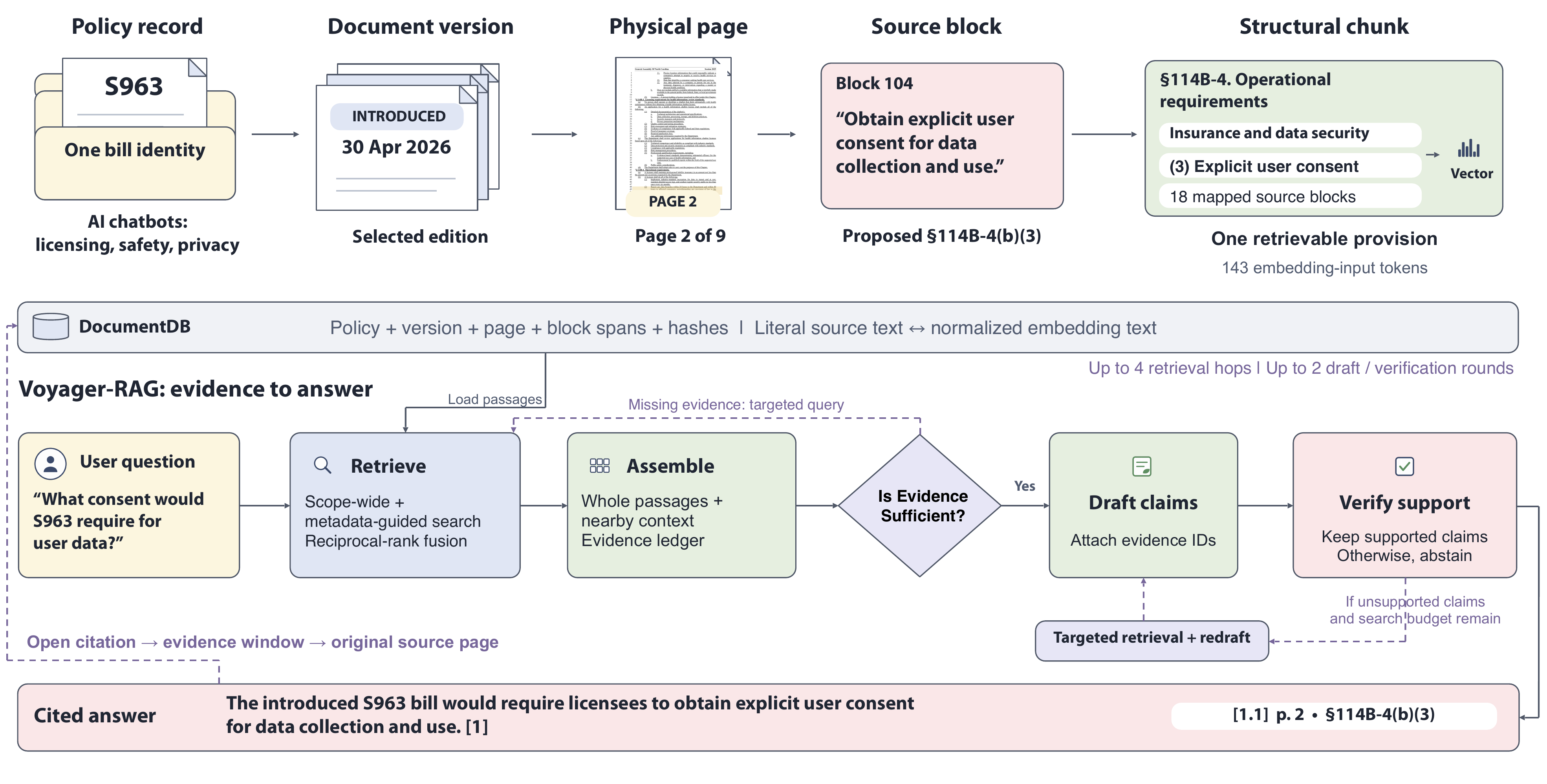}
    \caption{\textbf{How Voyager-RAG grounds an answer in versioned policy text.} \textbf{Top:} North Carolina Senate Bill 963 is traced from its policy record to the introduced version, page 2, a source block, and finally the passage containing the explicit user consent provision in proposed Section 114B-4(b)(3). Each passage records its document metadata, page and block references, character span, and source hash, so any citation can be traced to its exact location in the source. \textbf{Bottom:} Voyager-RAG answers a question about consent requirements. Scoped retrieval with reciprocal rank fusion finds relevant passages, which are combined with surrounding context into an evidence ledger. The controller checks that the evidence is sufficient before the model drafts claims with citation identifiers. The verifier checks each claim against its cited passages and requests targeted retrieval when support is missing. The final answer links the consent requirement to its source provision in the original document.}    \Description{Diagram tracing a policy record through its document version, physical page, source block, and structural passage, followed by retrieval, evidence checking, and a cited answer.}
    \label{fig:representation}
\end{figure*}

\paragraph{Approach.}
We represent governance knowledge at the level of source passages and their provenance (Figure~\ref{fig:representation}). A policy record is a catalogue entry such as one bill, and a document version is one indexed source for it, such as its introduced or enrolled text, with a canonical catalogue resolving category, jurisdiction, bill, and version metadata.

\paragraph{Structure-aware chunking.}
Semantic chunking adds cost without consistent gains~\cite{qu2025semantic}, so passages follow the document's own structure. Line wraps are grouped into paragraphs and provisions, and headings, list introductions, nested clauses, and attached qualifications stay together, while printed headers and page counters are recorded separately. Prose longer than an estimated 700 tokens is split at sentence boundaries. When a list continues into a new passage, the passage repeats the list introduction, up to 140 tokens and marked as repeated, so each clause keeps its scope. Tables, slides, worksheets, and unsplittable sentences stay whole and are flagged when oversized, and embedding enforces the 8,191-token input limit without silent truncation. The resulting 812,537 passages have a median length of 58 tokens and a 95th percentile of 353, and are embedded with \texttt{text-embedding-3-small} as 1,536-dimensional vectors.

\paragraph{Passage text, provenance, and manifests.}
Each passage stores its literal text and a normalized retrieval text. Normalization removes sequential margin numbers, joins line wraps, and repairs a split word only when the document or a frozen source-derived lexicon supports the joined spelling, and every operation records its offsets and replacement, so the normalized text can be reproduced from the source. Each passage records six provenance fields: document version, source hash, physical pages, source blocks, character spans, and position. A manifest binds each index to its source hashes, chunker version, model configuration, and artifact checksums, so citations stay resolvable as the corpus grows.

\paragraph{Reading aids.}
Reading aids follow the same rule. Summaries, mind maps, and key concepts are generated ahead of time from packets of at most 40 pages that combine the PDF with page images rendered at 180 DPI, with hierarchical note reduction (fan-in of eight) for longer documents. A generation identity binds each artifact to its document version, PDF hash, prompt, schema, model configuration, and producing code, and key concepts link to an exact passage only when their quoted text occurs once on the source page. Reading aids are generated with the Azure \texttt{gpt-6.1-sol} deployment. Shared page-attributed notes feed an eight-section summary, a hierarchical mind map, and up to six key concepts, each checkpointed separately, and raw responses are stored before parsing. Key concepts mark their source location as approximate when a quotation does not occur uniquely on its source page.

In the running example, clause (b)(3) and the list introduction it depends on become one passage, traceable to block 104 on page 2 of the introduced edition and to its character span.

\subsection{Voyager-RAG: Bounded Agentic Retrieval with Statement Verification}
\label{sec:rag-system}

\paragraph{Challenge.}
A single search for the question can return a passage that mentions consent and miss clause (b)(3), whose list introduction carries the duty. A model given the wrong passage can then state the requirement too broadly and cite a passage that does not support the claim~\cite{gao2023enabling}. In questions that compare bills, one long document can also fill the evidence window and crowd out the others.

\paragraph{Approach.}
A language-model controller decides when to search again, within limits it cannot change, and one \texttt{gpt-6-luna} deployment serves as controller, generator, and verifier. The catalogue resolves each question's scope, either the Companion's selected version or the collections chosen in AI Chat, and applies it inside every vector query. After each round, the controller judges whether the evidence answers the question and may issue a follow-up query. The generator returns statements with evidence identifiers, and a separate model call checks each statement against its cited passages, admits only supported statements, and requests another search when support is missing, within a shared cap of four rounds. In the running example, if the first round misses clause (b)(3), the controller issues a targeted query for it, and the verifier admits the consent statement only if (b)(3) supports it. These checks are model judgments and do not establish legal correctness.

\paragraph{Retrieval parameters.}
Passages are indexed with DiskANN~\cite{jayaram2019diskann} (\texttt{maxDegree}=64, \texttt{lBuild}=128, \texttt{lSearch}=512), and scopes of at most 4,096 passages use exact search. Each round retrieves 15 candidates by dense search over the scope and 15 more from preferred documents, and combines the lists $\mathcal{L}$ with reciprocal rank fusion~\cite{cormack2009reciprocal},
\begin{equation}
\operatorname{RRF}(c)=\sum_{L\in\mathcal{L},\,c\in L}\frac{1}{60+\operatorname{rank}_{L}(c)}.
\label{eq:rag-rrf}
\end{equation}

For questions that name up to four documents, each document receives its own bounded search, and the top eight passages are allocated across documents in round-robin order, ranking substantive body text above printing locators and unseen passages above repeated ones. Comparisons between two to four numbered provisions receive separate clause queries. Passages stay whole and in source order. A non-table passage shorter than 100 tokens is expanded with up to two neighbors on each side, targeting 200 tokens within a limit of 700, and expansion never crosses a version boundary or a missing page. Whole windows are admitted under an evidence budget of 10,000 tokens.

Dense retrieval scores passages by cosine similarity between query and passage embeddings, and when more than one jurisdictional index applies, the indexes are searched concurrently and merged by score. The controller, generator, and verifier run with reasoning effort set to \texttt{none}.

\paragraph{Revision, review, and traces.}
One bounded revision can repair an omission or wording problem, and if it fails, verified statements are kept and unsupported ones are dropped. In Companion discussions, the reviewer also checks that interpretations follow from the cited provisions and that proposed changes are labeled as proposals. Execution traces record queries, evidence selection, support decisions, and serving identities. Section~\ref{sec:implementation} describes the hosting infrastructure, and Appendix~\ref{app:prompts} gives the prompts.

\subsection{Implementation and Infrastructure}
\label{sec:implementation}

AI-Voyager uses a React frontend and Django APIs hosted on an Azure application VM with 48 vCPUs and 192\,GiB of RAM. Nginx terminates public HTTPS, serves frontend assets, and proxies API requests to Gunicorn, and Tailscale provides private administrative access. PostgreSQL~16 runs as a separate Azure Database for PostgreSQL Flexible Server, connected through private endpoints, private DNS, and hostname-verified TLS. Application traffic uses PgBouncer transaction pooling, while migrations, study provisioning, and cursor-dependent reporting use direct connections. PostgreSQL stores accounts, conversations, annotations, drafts, version history, study configuration, consent, questionnaire answers, and activity records. Each study has its own database and role, and managed backup and recovery operate independently of the application VM. DocumentDB stores the shared governance corpus, source and provenance records, vector indexes, and saved reading aids, and GridFS holds large source files. Redis supports bounded caches, coordination, and cancellation. Cached corpus reads are isolated by source and release identity, and personal records remain scoped to an account or study. At startup, workers validate serving manifests and load catalogue metadata. Azure OpenAI Foundry supplies embedding model \texttt{text-embedding-3-small} and OpenAI language-model API calls over HTTPS, and Azure Communication Services Email handles account verification and password recovery. PDF extraction and reading-aid generation run outside the request path.

\paragraph{Queued execution and recovery.}
Study model operations run as durable PostgreSQL jobs with up to four isolated worker processes per study. Admission allows one running job per participant and one editing job at a time, reserving capacity for interactive work, and queue wait is measured separately from execution. Interactive jobs have a 600-second execution deadline and editing jobs a 2,100-second deadline. Cancellation and recovery from failed workers keep completed outputs without automatically replaying interrupted requests. Versioned releases and source manifests bind runtime code to accepted corpus identities, and a rollback retains newer application records.

\section{Governance Benchmark}
\label{sec:benchmark}

The benchmark contains 360 questions in four categories: factual, temporal, comparative, and multi-hop. Each category has 30 questions about one, two, and four bills, following benchmarks of practical and multi-document information needs~\cite{kwiatkowski2019natural, tang2024multihop, krishna2025fact}. In 120 questions, the system must discover the relevant bills itself.

\paragraph{Snapshot and editions.}
The benchmark uses a frozen snapshot of 2,166 bill groups and 3,501 document versions: 2,111 groups and 3,446 versions from the U.S. federal and state collections, 29 groups and versions of national legislation, and 26 groups and versions of city policy. Each official edition keeps its own version identifier and source hash. Questions use one representative edition per group, preferring final or enrolled text, then introduced text, then the latest draft. Evidence records the document and version identifiers, page, exact text span, and page-content hash. Catalogue descriptions may guide discovery, but every legislative claim must be supported by source text.

\paragraph{Domain taxonomy.}
The taxonomy uses the 21 parent domains from the Comparative Agendas Project~\cite{jones2025policyagendas} and corpus-specific subdomains derived following the grounds of TopicGPT~\cite{pham2024topicgpt}. Of the 3,501 versions, 3,456 received a primary domain and 45 remained uncertain, and versions may also receive secondary domains. Independent source inspection covered 1,094 versions, including a stratified audit of 100 records (40 federal, 40 state, 10 national legislation, and 10 city). These counts describe audit coverage and do not establish classification accuracy.

\paragraph{Eligibility.}
Of the 2,166 groups, 5 lack an eligible representative edition and 23 have an uncertain primary domain, leaving 2,138. We held out 57 groups used in the construction pilot, attempted pilot source sets, or an earlier 22-question diagnostic benchmark, leaving 2,081 groups. Each group counts once toward its primary domain regardless of how many editions it has, and every domain has at least 12 eligible groups.

\paragraph{Allocation.}
Domain quotas use square-root weighting with a per-domain minimum:
\begin{equation}
\widetilde{q}_d = m + (Q-mD)
\frac{\sqrt{N_d}}{\sum_{j=1}^{D}\sqrt{N_j}},
\qquad Q=360,\;m=6,
\label{eq:domain-allocation}
\end{equation}
where $N_d$ is the number of eligible groups in domain $d$ and $D=21$. The minimum protects small domains, and the square root reflects corpus size without letting large domains dominate. The minimum allocates 126 questions and the remaining 234 are distributed by $\sqrt{N_d}$. Flooring gives 351, and largest-remainder rounding allocates the last nine (Table~\ref{tab:internal-domain-allocation}). The allocation also requires at least 30 questions involving national legislation, 30 involving city policy, and 36 involving documents over 80 pages, including six over 1,000 pages, and each group appears in at most four questions.

\paragraph{Question construction and review.}
GPT-6.1-Sol generated each question with a reference answer and evidence spans. Source-aware model review accepted, corrected, or rejected each case. The questions are synthetic and were not sampled from user traffic. For each question, source passages support a concise reference answer and usually one to three points that a correct answer must include. Evidence for long documents keeps the required definitions, exceptions, and cross-references, along with any limits on coverage. Review checks each question's naturalness, source sufficiency, edition identity, reasoning requirements, and scoring rubric. Multi-hop questions record the dependency structure that connects the required provisions to the conclusion. In discovery questions, an alternative bill counts only after verification that it meets the requested conditions and contributes a supported finding. The benchmark was frozen before any system outputs were assessed, and rejected or repaired cases keep their audit records.
\begin{table}[t]
\centering
\normalsize
\setlength{\tabcolsep}{2pt}
\renewcommand{\arraystretch}{1.08}
\caption{Sampling population and question allocation by primary domain. $N_d$ counts eligible bill groups after holdouts, and $q_d$ is the quota from Equation~\ref{eq:domain-allocation}. CAP codes keep their original numbering.}
\label{tab:internal-domain-allocation}
\begin{tabularx}{\columnwidth}{
    @{}r>{\raggedright\arraybackslash}Xrr@{}
}
\toprule
CAP & Primary policy domain & $N_d$ & $q_d$ \\
\midrule
1  & Macroeconomics & 50 & 15 \\
2  & Civil rights and civil liberties & 56 & 15 \\
3  & Health & 197 & 23 \\
4  & Agriculture & 34 & 13 \\
5  & Labor and employment & 100 & 18 \\
6  & Education & 148 & 21 \\
7  & Environment & 44 & 14 \\
8  & Energy & 46 & 14 \\
9  & Immigration & 30 & 13 \\
10 & Transportation & 34 & 13 \\
12 & Law, crime and family issues & 157 & 22 \\
13 & Social welfare & 12 & 10 \\
14 & Housing and community development & 26 & 12 \\
15 & Domestic commerce & 249 & 25 \\
16 & Defense & 148 & 21 \\
17 & Space, science, technology and communications & 334 & 29 \\
18 & Foreign trade & 74 & 17 \\
19 & International affairs and foreign aid & 81 & 17 \\
20 & Government operations & 213 & 24 \\
21 & Public lands and water management & 23 & 12 \\
23 & Culture & 25 & 12 \\
\midrule
& \textbf{Total} & \textbf{2,081} & \textbf{360} \\
\bottomrule
\end{tabularx}
\end{table}
\paragraph{Human validation.}
Two independent auditors rated system answers to 100 benchmark questions for correctness, groundedness, and citation support on an ordinal scale from 0 to 4, and we measured agreement with quadratic-weighted Cohen's $\kappa$. Three questions that the content filter blocked received \texttt{NA} in both passes and were excluded, leaving 97 paired observations per metric. Agreement was $\kappa_w=0.969$ for correctness and groundedness and $0.979$ for citation support. Exact agreement was 94.8\% (92 of 97) for each metric, every disagreement differed by one scale point, and disagreements were resolved by consensus.

\section{System Evaluation}
\label{sec:evaluation}

We evaluate evidence retrieval in both open-document and fixed-document settings.

\subsection{Experimental Setup}
\label{sec:eval-setup}

\paragraph{Governance benchmark.}
The governance benchmark in Section~\ref{sec:benchmark} evaluates retrieval when relevant bills and editions must first be discovered.

\paragraph{LegalBench-RAG.}
\label{sec:legalbench}
We use the released corpus, questions, and reference spans of LegalBench-RAG-mini: 776 questions over 72 contracts and privacy policies with 1,307 gold snippets~\cite{pipitone2024legalbench}, evaluated for retrieval alone through the same retrieval interface (Appendix~\ref{app:legalbench}).

\paragraph{Baselines and ablations.}
\label{sec:rag-ablation}
BM25~\cite{robertson2009bm25}, single-pass dense retrieval with the same embeddings, and a hybrid of the two through reciprocal rank fusion run over the same passages and scopes, without iteration, per-bill targeting, context expansion, or verification. Four ablations each disable one of these components of Voyager-RAG (Section~\ref{sec:eval-metrics}).

\subsection{Metrics and Statistical Procedures}
\label{sec:eval-metrics}

\paragraph{Evaluation units and metrics.}
Neither evaluation gives the system gold document identifiers. Each governance question ran once in a fresh conversation. Retrieved passages are pooled across rounds using reciprocal rank fusion for measurement only. A passage counts as gold when it overlaps a reference span in the same document version. We report macro precision and recall, averaged over per-question ratios, at $k=15$ in the main text and at $k\in\{2,4,8,15\}$ in Appendix~\ref{app:master}. Precision on discovery questions is conservative, since valid sources can lie outside the reference set.

\paragraph{Ablations.}
Without iterative retrieval, Voyager-RAG runs one round with no follow-up searches, including those the verifier requests. Without claim verification, it performs no semantic support checking or verifier-requested search, and citation identifier validation is kept.

\paragraph{Pooled ranking.}
For question $i$, the pooled score $s_i(c)$ of passage $c$ is calculated using RRF in Equation~\ref{eq:rag-rrf} and the top-$k$ set is calculated as $R_i^{(k)}=\operatorname{Top}_k(s_i)$.

\paragraph{Scoring details.}
In the pooled ranking $R_i^{(k)}$, each round has equal weight, and ties are broken by best rank, first occurrence, and passage identifier. Gold overlap requires at least one shared non-whitespace character, and precision is zero for an empty return. Retrieved counts can fall below the nominal cutoff when fewer passages are available.

\paragraph{Significance tests.}
We used 10,000 bootstrap resamples (seed \texttt{20261102}), stratified by question type $\times$ bill count with fixed stratum sizes, and shared the sampled question identifiers across configurations. At $k=15$, we compared each baseline and ablation with full Voyager-RAG using two-sided centered paired-bootstrap tests on macro precision and macro recall, with Holm correction across the 14 comparisons. An adjusted $p$ of 0.0014 is the smallest value this procedure can return.

\paragraph{Latency.}
Latency also depends on call duration, retrieval execution, and provider response times, so model-call counts do not determine it alone.

\paragraph{Filtered questions.}
Counting the four questions blocked by the content filter as failures lowers macro precision from 67.5\% to 66.8\% and macro recall from 83.0\% to 82.1\% at $k=15$.

\subsection{Results}
\label{sec:eval-results}

\subsubsection{Retrieval finds nearly all evidence when documents are fixed}
On LegalBench-RAG, macro recall rises from 77.3\% at $k=2$ to 97.2\% at $k=8$ and 99.1\% at $k=15$, and at $k=15$ every subset lies between 98.7\% and 99.5\% (Appendix~\ref{app:master}). Precision at $k=15$ reaches 11.1\%, against a ceiling of 11.2\% set by the number of gold snippets per question (Appendix~\ref{app:legalbench}).

\subsubsection{Discovery and cross-document evidence make the open setting harder}
On the governance benchmark, macro recall rises from 25.8\% at $k=2$ to 83.0\% at $k=15$, while macro precision falls from 88.8\% to 67.5\% (Table~\ref{tab:eval-summary}). Factual questions have the highest precision (73.5\%) and temporal questions the highest recall (89.9\%; Appendix~\ref{app:master}). The settings differ in corpus and evidence definitions, and the lower coverage in the open setting is consistent with the added steps of document discovery, edition selection, and cross-document evidence. Hybrid retrieval is the strongest baseline (60.3\% precision, 80.5\% recall), and Voyager-RAG exceeds it by 7.2 points in precision and 2.5 in recall.

\subsubsection{Iterative retrieval and statement verification account for most of the coverage}
Removing iterative retrieval lowers macro recall by 9.3 points to 73.7\%, while cutting model calls from 8.6 to 4.2 and p95 latency from 15.1 to 4.6 seconds. Removing claim verification lowers recall by 6.5 points, because the searches the verifier requests recover evidence the first rounds missed. Removing context expansion lowers it by 2.4 points at the same number of model calls, and removing per-bill targeting by 0.4. In the paired tests on macro metrics, the first three recall reductions are significant (Holm-adjusted $p=0.0014$, the smallest value 10,000 resamples allow), the reduction without per-bill targeting is not, and all four ablations lower precision significantly. Without iteration, Voyager-RAG falls below single-pass dense retrieval, because its first round applies document targeting, rank fusion, and source balancing, which can displace relevant passages that only later rounds recover.

\begin{table}[!tpb]
\centering
\setlength{\tabcolsep}{4pt}
\renewcommand{\arraystretch}{1.13}
\caption{Retrieval results at $k=15$. Q counts completed questions. Macro precision and recall are percentages. p95 Latency and model calls are per question. Baselines and ablations use the same 356 questions. Appendix~\ref{app:master} gives results by question type and at other cutoffs.}
\label{tab:eval-summary}
\begin{adjustbox}{max width=\linewidth}
\begin{tabular}{@{}lrrrrr@{}}
\toprule
Evaluation set / configuration & Q
& $P_{\mathrm{macro}}$
& $R_{\mathrm{macro}}$
& \shortstack{p95\\Latency (s)}
& \shortstack{Model calls\\per question} \\
\midrule
\multicolumn{6}{@{}l}{\textbf{LegalBench-RAG-mini}} \\
Voyager-RAG (overall) & 776 & 11.1 & 99.1 & 7.0 & 4.8 \\
\addlinespace[2pt]
\multicolumn{6}{@{}l}{\textbf{Governance benchmark: \emph{Voyager-RAG}}} \\
Voyager-RAG (overall) & 356 & 67.5 & 83.0 & 15.1 & 8.6 \\
\addlinespace[2pt]
\multicolumn{6}{@{}l}{\textbf{Governance benchmark: \emph{reference baselines}}} \\
BM25 & 356 & 51.4 & 72.9 & 5.6 & 1.01 \\
Single-pass dense & 356 & 58.9 & 78.5 & 18.3 & 2.01 \\
Hybrid (BM25 + dense) & 356 & 60.3 & 80.5 & 18.1 & 2.01 \\
\addlinespace[2pt]
\multicolumn{6}{@{}l}{\textbf{Governance benchmark: \emph{ablations}}} \\
Without iterative retrieval & 356 & 56.2 & 73.7 & 4.6 & 4.2 \\
Without per-bill targeting & 356 & 64.1 & 82.6 & 14.4 & 8.8 \\
Without context expansion & 356 & 62.5 & 80.6 & 11.7 & 8.6 \\
Without claim verification & 356 & 58.3 & 76.5 & 9.3 & 7.1 \\
\bottomrule
\end{tabular}
\end{adjustbox}
\end{table}
 
\subsection{Interpretation and Boundaries}
\label{sec:technical-summary}

\paragraph{Interpreting precision and recall.}
At $k=15$, Voyager-RAG achieves $83.0\%$ macro recall, averaged across $356$ completed governance questions. Answer accuracy requires separate assessment. Prior work finds that conventional retrieval metrics can correlate weakly with downstream RAG performance~\cite{salemi2024evaluating}, and evaluates answer correctness and citation support as distinct outcomes~\cite{gao2023enabling}. Precision and recall also depend on the amount of reference evidence. For gold and retrieved sets expressed in the same distinct passage units, let $g_i$ denote the number of gold passages for question $i$ and $n_i=|R_i^{(k)}|\leq k$ the number retrieved. For questions with $n_i>0$, their cardinalities imply $P_i@k \leq \min(1,g_i/n_i)$ and $R_i@k \leq \min(1,n_i/g_i)$. Thus, questions with more gold passages than retrieval positions cannot achieve complete recall at that cutoff. LegalBench-RAG averages $1.68$ gold units per question, giving a precision ceiling of $1307/(776\times15)=11.2\%$ when every question returns 15 passages. The observed $11.1\%$ precision is close to this ceiling. The governance benchmark averages $5347/356=15.02$ gold units per question. These averages contextualize retrieval depth within each benchmark's annotation scheme, although exact ceilings depend on the per-question distribution. Macro recall assigns equal weight to every question, regardless of its reference-set size. Across our evaluated cutoffs, increasing $k$ raises recall while lowering precision as additional unmatched passages enter the retrieved set. This tradeoff concerns evidence coverage and context selectivity. Published experiments show that the number and composition of retrieved passages also affect generated-answer quality~\cite{cuconasu2024power}. Our cutoff analysis evaluates pooled retrieval rankings without regenerating answers, so it does not measure changes in answer accuracy.

\paragraph{Boundaries.}
The provider's content filter blocked four questions about public law, which a public deployment must handle. Section~\ref{sec:eval-metrics} reports the effect of counting these as failures. These results describe retrieval and model-judged support, and answer accuracy needs separate assessment~\cite{salemi2024evaluating, gao2023enabling}.

\section{Human-AI Evaluation}
\label{sec:study}

AI legislation often regulates harms that have not yet materialized, so making sense of it means working out what it could mean for different people. The technical evaluation cannot show whether a perspective helps with this or how people judge what they are told. We therefore studied how people use AI-Voyager to interpret real AI bills, examining whether lenses change which provisions and stakeholders people attend to (DG3), whether verifiable citations lead people to check what they are told (DG4), and what people propose after interpreting a bill with AI assistance (DG5). In terms of the grammar, the study covers Read, Ask, Reframe, and Verify, examines Revise through verbal proposals, and exercises Find only for a known document.

\subsection{Setup}
\label{sec:study-setup}

\paragraph{Participants.}
We recruited 24 participants (16 men and 8 women), aged 18 to 24, through email outreach and word of mouth. All reported fluency in English. On 7-point scales, most reported frequent use of generative AI over the preceding three months (median 6.5, range 2--7), less engagement with formal policy documents over the past year (median 3, range 1--6), and little familiarity with either bill (median 1, range 1--2). Participation in our study was voluntary and unpaid, and our institution's Institutional Review Board approved the study.

\paragraph{Bills.}
Participants worked with two AI bills at different stages of the legislative process. The hiring bill, New York Senate Bill S8831~\cite{ny_senate_s8831_2026}, is an enacted law on the use of automated employment decision tools by public entities. The chatbot bill, North Carolina Senate Bill 963~\cite{nc_senate_sb963_2026}, is a proposed bill on the licensing, safety, and privacy of AI chatbots that handle health information.

\paragraph{Procedure.}
Prior to the main study, we conducted six pilot sessions (P3, P4, P10, P12, P15, and P19) to refine the protocol. These are part of the 24 participants recruited. Based on these sessions, we shifted from written responses to our probes to verbal reflections, extended the AI exploration period, and expanded the closing interview.

Each participant completed two tasks in a remote within-subjects session of about 55 minutes. In one task, participants used General Analysis alone. In the other, they could also choose among the five lenses (Civilian, Policymaker, AI developer, Philosopher and Alignment Agent) and were asked to try several. The model, documents, and citations were the same in both, and the order of conditions and bills was counterbalanced across four groups of six (Appendix~\ref{app:study}). After a demonstration that included opening a citation, participants imagined volunteering for a community organization that wanted a plain-language explanation of each bill. They found the bill in the Bill Library, read it for three minutes without AI, and explored it with the AI Companion for nine minutes while thinking aloud. With the AI hidden, they had seven minutes to answer five questions aloud about the bill's purpose, provisions, stakeholders, differing perspectives, and changes or clarifications. A ten-minute closing interview asked about preference, stakeholders, trust, the lenses, and whether the system shaped their own judgment. We collected screen and audio recordings, interaction logs, and interviews. Appendix~\ref{app:study} gives the instructions for each condition.

\paragraph{Analysis.}
One researcher coded the transcripts inductively for how participants selected and used lenses, trusted AI-generated information and checked sources, proposed changes or clarifications, and used AI to understand legislation and form their own judgments. A second researcher, who had conducted every session, reviewed the resulting themes, and the two refined them through discussion until they agreed (codebook in Appendix~\ref{app:study}). We also validated the answers participants received to ``What is this bill about?'' against each bill's PDF and citations (Appendix~\ref{app:study-validation}). The study surfaces how people used the system and the design questions this raises.

\subsection{Findings}
\label{sec:study-findings}
Appendix~\ref{app:study-findings} gives additional quotes for each finding.

\subsubsection{Participants asked how provisions would apply in practice (DG5)}
Asked what they would change or clarify, participants mostly asked how provisions would be carried out, and few objected to the bills on ideological grounds. P2 noted that the chatbot bill would grant licenses but did not ``explain [...] how that process is gonna go.'' P24 asked what would happen to non-health information after the bill's 60-day retention period: ``Is it deleted after the 60 days? Is it kept longer than 60 days? Where does it go?'' P18 saw nothing in the hiring bill that ``makes sure that this is being followed.'' Others asked about scope. P14 asked which industries and employees the hiring bill covered, P8 asked which health-related chatbot interactions required licensing, and P17 asked what the hiring bill meant by AI: ``Is that machine learning? Is that ChatGPT? What are we talking about?'' These questions sat where a bill's definitions meet specific people and situations. For DG5, reading the exact text with AI assistance led participants to precise, checkable clarifications, the kind of contribution that Revise is built to carry forward.

\begin{table*}[t]
\centering
\renewcommand{\arraystretch}{1.5}
\setlength{\tabcolsep}{3pt}
\newcolumntype{Y}{>{\raggedright\arraybackslash}X}
\caption{A design space for human-AI civic reasoning,
derived from building, evaluating, and studying AI-Voyager.}
\label{tab:design-space}
\begin{tabularx}{\linewidth}{@{}p{1.5cm}p{2.5cm}YYY@{}}
\toprule
\textbf{Layer} &
\textbf{Dimension} &
\textbf{Tension} &
\textbf{AI-Voyager's choice} &
\textbf{Evidence from this work} \\
\midrule
Technical &
Unit of evidence &
Readable summaries vs.\ faithful source text &
Source passages only, with character-level provenance &
Reading aids kept apart from evidence
(\S\ref{sec:representation}) \\
&
Representation over time &
Corpus-level structure vs.\ qualifications and editions kept intact &
Versioned passages bound by manifests &
3,653 versions with resolvable citations
(\S\ref{sec:corpus}) \\
&
AI agency &
Coverage vs.\ cost and predictability &
Bounded agentic retrieval with statement verification &
83.0\% vs.\ 73.7\% macro recall with and without iteration,
at twice the calls (\S\ref{sec:eval-results}) \\
\midrule
Human-AI &
Interpretation &
Breadth of perspectives vs.\ steering the reading &
Five lenses over one shared evidence base &
Lenses saved time; General Analysis left room for own thinking
(\S\ref{sec:study-findings}) \\
&
Verification burden &
Convenience vs.\ scrutiny &
Model checks each statement; users can open each citation &
Citations read as grounding cues, rarely used to verify
(\S\ref{sec:study-findings}) \\
\midrule
Societal &
Participation &
Lower effort vs.\ meaningful influence &
Reading, drafting, and public discussion &
Participants asked how provisions apply; no institutional channel yet
(\S\ref{sec:study-findings}) \\
&
Infrastructure &
Model capability vs.\ independence and cost &
Commercial models with recorded serving identities &
4 questions blocked, 2 documents refused;
8.6 calls per question (\S\ref{sec:evaluation}) \\
&
Coverage &
Breadth vs.\ depth across jurisdictions &
Corpus collected from scratch, plus uploads from any jurisdiction &
Corpus mirrors where AI legislation exists; uploads extend it
(\S\ref{sec:corpus}) \\
\bottomrule
\end{tabularx}
\end{table*}

\subsubsection{Participants valued lenses for broader, faster exploration and General Analysis for independent thinking (DG3)}
Most participants preferred the lenses, which helped them consider more issues without finding each one themselves. P23 said, ``I think I could still get to most of these points [...] if I had enough time, but it cuts down on the time.'' Participants also valued General Analysis for more deliberate exploration. P22 said that ``not having the perspectives allowed me as a human to think a little bit more,'' and P16 cautioned that predefined perspectives could put issues into ``buckets'' and ``restrict thinking.'' For DG3, the lenses widened what participants considered at the cost of some self-directed engagement.

\subsubsection{Participants chose lenses for the kind of explanation they wanted as well as the viewpoint (DG3)}
Several participants began with General Analysis or Civilian to understand the bill. P13 wanted ``a general consensus of what this text is about'' before changing perspectives, and P24 chose Civilian because it ``put it in layman's terms.'' Others followed identity or role, as when P5 chose Civilian ``because I am a civilian.'' P23 chose AI Developer because ``this perspective will be very different than mine.'' Because participants were asked to try several lenses, we treat these patterns as descriptive. For DG3, participants used lenses as reading tools for explanation style and depth, in addition to stakeholder viewpoints.

\subsubsection{Participants described citations both as cues that AI responses were grounded in the legislation and as tools for navigating the underlying text (DG4)}
For some participants, the presence of citations raised trust. P1 said that ``seeing that it did cite its sources and being able to click on it [...] made me trust it more,'' and P13 ``assumed that it was [...] pulling it correctly.'' Others used citations to find passages, as P21 did to ``see exactly where from the [...] document it pulled from.'' Having citations did not mean using them to check. P20 described citations as a way to ``cross-check and cross-verify the sources'' and then said, ``I have not checked the sources here.'' For DG4, Verify was one click away on every answer, yet trust came mainly from the availability of citations.

\subsubsection{Participants described using AI to understand and explore the legislation while considering policy judgments their own responsibility (DG5)}
P13 said that ``the AI helped me a lot in gathering information [...] but at the end of the day, I felt like I was the one who was making my own thoughts.'' Suggested follow-up questions still steered what participants explored. P24 followed a suggested question about the 60-day retention provision, concluding, ``It does not require deletion after the 60 days, so what does happen after the 60 days?'' This became one of P24's proposed clarifications. P2 reflected that if they were ``actually involved in [...] this decision-making,'' where their ``voice has more impact,'' they would ``want to be more intentional with the questions that I ask.'' For DG5, participants viewed their final judgments as their own, while the system's suggested questions could still shape which issues they considered along the way, suggesting a potential tradeoff in agency.

\section{Discussion}
\label{sec:discussion}

We turn the study's patterns into design considerations and generalize them into a design space for any system that mediates public access to law.

\subsection{Design Considerations}
\label{sec:disc-grammar}

Participants built their reasoning mainly from Read, Ask, and Reframe. They trusted citations without opening them, let suggested questions steer some of what they explored, and produced their most specific contributions through Revise, performed aloud.  Four considerations follow, each tied to one operation.

\textbf{C1. Build Verify into Ask.} Showing the cited passage beside each claim, and asking readers to confirm a key provision before they draft, makes checking part of reading, since overreliance falls when checking is cheap~\cite{bucinca2021trust, vasconcelos2023overreliance}.

\textbf{C2. Keep the reader's own framing one step away.} Since lenses saved time while General Analysis left more room for participants' own thinking, Reframe should keep a neutral mode available, show which provisions a lens left out, and describe each lens as an analysis that does not report the views of the group it names.

\textbf{C3. Make suggested questions optional prompts.} Suggestions led participants to issues they later raised as clarifications. Ask should present them as options among several and leave room for the reader's own questions.

\textbf{C4. Carry clarifications into drafting.} Revise should bring the clarifications people raise into Edit Bill with their sources attached.

\subsection{A Design Space for Human-AI Civic Reasoning}
\label{sec:disc-design-space}

Table~\ref{tab:design-space} generalizes these choices into eight dimensions across technical, human-AI, and societal layers. The lower macro recall in the open setting (83.0\%) than on fixed documents (99.1\%) shows why such systems must be evaluated where documents must first be found. Lenses sped up exploration while General Analysis supported more self-directed thinking, and model checks on each statement left the reader's own verification undone. Participants located underspecification, a role for the public that differs from stating preferences~\cite{lee2019webuildai}, and such contributions gain weight when they reach the institutions that draft laws~\cite{r2013regulation, aitamurto2016crowdsourced}. Coverage is where users hold the most agency: the curated corpus mirrors where AI legislation exists, and uploads let a resident of any country bring the text that governs them.

\subsection{Implications for Alignment and Measurement}
\label{sec:disc-alignment}
\label{sec:disc-measurement}

For alignment, AI-Voyager complements work on public input into model principles~\cite{bai2022constitutional, huang2024collective} and on how people adapt to AI systems~\cite{shen2024towards}: the public examines and revises the rules that govern how AI is deployed. Participants' requests suggest that laws and model specifications share a problem, since written principles leave their application unclear in specific cases, and because lenses shaped what readers explored, evaluations of pluralistic systems~\cite{sorensen2024roadmap} should include their effect on users. For measurement, retrieval should be evaluated where documents must be discovered, and grammar operations logged, since citation counts and support checks do not show whether anyone read the source.

\section{Limitations and Future Work}
\label{sec:disc-limitations}

The corpus will grow as more countries legislate on AI, and a datasheet will document its sources and inclusion criteria~\cite{gebru2021datasheets}. The benchmark questions were constructed with language models and checked against sources. Future versions can add questions from public use and references verified by people, and legal experts can assess correctness beyond retrieval and model-judged support. The user study involved 24 participants aged 18 to 24 in relatively short sessions, limiting the generalizability of our findings. Future work could examine longer-term engagement with the system and its use among more diverse populations. Edit Bill, Launchpad, document upload, and the design considerations need studies of their own, and the public launch will show who contributes over time, how representative contributions are, and whether proposals reach the institutions that draft laws.

\section{Conclusion}
\label{sec:conclusion}

In this paper, we introduced AI-Voyager, a public-interest system for human-AI civic reasoning, in which people and AI work together to understand, contest, and shape the rules that govern AI. We built an architecture that ties every claim to one passage in one edition across 3,653 governance documents, and found on a new benchmark that retrieval is harder when documents must first be found. In a study with 24 participants new to two AI bills, we found that people asked how provisions would apply, while the system's lenses and suggested questions shaped what they explored and its citations shaped what they trusted. Systems like this should be designed and evaluated as participants in the reasoning they support. As AI-Voyager opens to the public, anyone, anywhere, can bring the rules that govern AI in their community, question them against their source, and propose how they should change.

\section*{Ethical Considerations and Adverse Impact Statement}
\label{sec:disc-risks}
While AI-Voyager enables people across the world to participate in shaping responsible AI designs, it also carries two risks, and we addressed each in its design. First, users may treat its answers as accurate statements of the law. AI-Voyager admits only statements that its verifier judges to be supported by the cited sources, keeps each source passage one click away, and says when support is unavailable, so users can check every claim against its source. Second, users may mistake model-generated perspectives for the voices of the groups they name. We label AI-Voyager's lenses as analytical viewpoints over shared evidence and publish the prompts that define them (Appendix~\ref{app:prompts}). Beyond these safeguards, the AI-Voyager team maintains human oversight of the system.

\section*{Generative AI Usage Statement}
AI-Voyager uses GPT-6-family models for retrieval control, answer generation, and verification, as described in the technical sections. Model-generated benchmark questions underwent source-aware review, and two auditors assessed system answers to a subset of benchmark questions. The authors used Claude Opus 5.5 for grammar checks and editing of author text, in accordance with conference guidelines.

\section*{Acknowledgments}
We thank the UNC AI Acceleration Program for providing compute resources, Atharv Chakole for help with data collection, and study participants for their valuable time and contributions.

\clearpage
\onecolumn

\bibliographystyle{ACM-Reference-Format}
\bibliography{references}

\clearpage

\appendix

\onecolumn

\section{LegalBench-RAG Details}
\label{app:legalbench}

This appendix provides the benchmark composition, retrieval setup, and evidence capacity analysis for LegalBench-RAG-mini. Table~\ref{tab:legalbench-composition} reports the distribution of questions and gold snippets across its four subsets. Table~\ref{tab:legalbench-capacity} summarizes the retrieval cutoffs and corresponding precision ceilings.

\paragraph{Comparison with the original paper.}
Pipitone and Houir Alami~\cite{pipitone2024legalbench} report 4.36\% precision and 26.30\% recall at $k=8$ for recursive character splitting without reranking on LegalBench-RAG-mini. Their character-level overlap metrics and different chunking scheme prevent a direct numerical comparison with our passage-level evidence scores.

\begin{table}[H]
\centering
\caption{Composition of LegalBench-RAG-mini. Gold/Q gives the minimum, mean, and maximum number of gold snippets per question.}
\label{tab:legalbench-composition}
\begin{tabular*}{0.90\linewidth}{@{}l@{\extracolsep{\fill}}rrrc@{}}
\toprule
Subset & Questions & Gold & Gold (\%) & Gold/Q \\
\midrule
ContractNLI & 194 & 276 & 21.1 & 1 / 1.42 / 3 \\
CUAD        & 194 & 272 & 20.8 & 1 / 1.40 / 7 \\
MAUD        & 194 & 306 & 23.4 & 1 / 1.58 / 7 \\
PrivacyQA   & 194 & 453 & 34.7 & 1 / 2.34 / 14 \\
\midrule
Overall     & 776 & 1,307 & 100.0 & 1 / 1.68 / 14 \\
\bottomrule
\end{tabular*}
\end{table}

\paragraph{Setup.}
The supplied text enters a separate adapter without OCR or Markdown cleanup that preserves file identities, hashes, and the offsets needed to map passages to benchmark evidence. Every annotation is checked against its source substring before indexing, and questions, gold evidence, and question-to-document mappings stay outside the searchable corpus.

\paragraph{Composition and precision ceilings.}
\label{app:legalbench-composition}
The 776 questions are split evenly across subsets, but reference evidence is not (Table~\ref{tab:legalbench-composition}). PrivacyQA contributes 34.7\% of the 1,307 gold snippets, averaging 2.34 per question and reaching 14 for one question. A question with more than $k$ gold snippets cannot reach full recall at cutoff $k$, and the number of gold snippets also caps precision. Table~\ref{tab:legalbench-capacity} gives, for each subset, the smallest evaluated cutoff at which every question can reach full recall and the highest precision possible at that cutoff. Across the benchmark, precision at $k=15$ cannot exceed 11.2\%, and Voyager-RAG reaches 11.1\%.

\begin{table}[H]
\centering
\caption{Retrieval capacity. Full-capacity $k$ is the smallest evaluated cutoff that can hold every gold snippet for any question in the subset. Max precision assumes all gold snippets are retrieved at that cutoff.}
\label{tab:legalbench-capacity}
\begin{tabular*}{0.90\linewidth}{@{}l@{\extracolsep{\fill}}rrr@{}}
\toprule
Subset & Max Gold/Q & Full-capacity $k$ & Max precision (\%) \\
\midrule
ContractNLI & 3  & 4  & 35.6 \\
CUAD        & 7  & 8  & 17.5 \\
MAUD        & 7  & 8  & 19.7 \\
PrivacyQA   & 14 & 15 & 15.6 \\
\midrule
Overall     & 14 & 15 & 11.2 \\
\bottomrule
\end{tabular*}
\end{table}

Table~\ref{tab:legalbench-capacity} shows how the number of gold snippets per question determines the retrieval capacity and attainable precision. Full-capacity $k$ is the smallest evaluated cutoff among $\{2,4,8,15\}$ that can accommodate all gold snippets for every question in a subset. ContractNLI requires only $k=4$, since no question has more than three gold snippets, while PrivacyQA requires $k=15$ to accommodate questions with up to 14 snippets. At full capacity, the maximum macro precision is the mean number of gold snippets per question divided by $k$, assuming exactly $k$ passages are retrieved. This explains why precision ceilings range from 35.6\% for ContractNLI to 15.6\% for PrivacyQA. Across all 776 questions, the maximum macro precision at $k=15$ is 11.2\%. Voyager-RAG achieves 11.1\%, approaching this ceiling while reaching 99.1\% macro recall.

\clearpage

\section{Governance Benchmark Complete Results}
\label{app:master}

Table~\ref{tab:rag-master} presents the complete retrieval results across both evaluation benchmarks. The table reports macro precision and recall at retrieval cutoffs $k=2,4,8,15$ for LegalBench-RAG-mini and the governance benchmark, including results by subset and question type. It also reports the reference baselines and component ablations at $k=15$, along with latency and model-call statistics.

\begingroup
\normalsize
\setlength{\tabcolsep}{3pt}
\renewcommand{\arraystretch}{1.15}
\begin{longtable}{@{}>{\raggedright\arraybackslash}p{0.29\linewidth}rrrrrrr@{}}
\caption{Detailed results of technical evaluation. The governance overall result in section B at $k=15$ is the full-system reference for baselines and ablations in section C.}
\label{tab:rag-master} \\
\toprule
\shortstack[l]{Subset / configuration} &
$k$ & Q & Gold & $P_M$ & $R_M$ &
\shortstack{p95 Latency (s)} &
\shortstack{Model calls per question} \\
\midrule
\endfirsthead
\toprule
\shortstack[l]{Subset / configuration} &
$k$ & Q & Gold & $P_M$ & $R_M$ &
\shortstack{p95\\Latency (s)} &
\shortstack{Model calls\\per question} \\
\midrule
\endhead
\midrule
\multicolumn{8}{r@{}}{Continued on next page} \\
\endfoot
\bottomrule
\endlastfoot
\multicolumn{8}{@{}l}{\textbf{A. LegalBench-RAG-mini: Voyager-RAG (776 questions)}} \\
\midrule
\textbf{Overall} & 2 & 776 & 1,307 & 62.0 & 77.3 & 7.0 & 4.8\\
ContractNLI & 2 & 194 & 276 & 55.9 & 80.8 & 6.5 & 4.6\\
CUAD & 2 & 194 & 272 & 53.6 & 78.8 & 6.9 & 4.8\\
MAUD & 2 & 194 & 306 & 61.6 & 81.1 & 7.0 & 4.9\\
PrivacyQA & 2 & 194 & 453 & 76.8 & 68.5 & 7.4 & 5.0\\
\addlinespace[2pt]
\textbf{Overall} & 4 & 776 & 1,307 & 37.5 & 90.9 & 7.0 & 4.8\\
ContractNLI & 4 & 194 & 276 & 31.3 & 89.4 & 6.5 & 4.6\\
CUAD & 4 & 194 & 272 & 31.8 & 92.6 & 6.9 & 4.8\\
MAUD & 4 & 194 & 306 & 35.8 & 92.8 & 7.0 & 4.9\\
PrivacyQA & 4 & 194 & 453 & 50.9 & 88.9 & 7.4 & 5.0\\
\addlinespace[2pt]
\textbf{Overall} & 8 & 776 & 1,307 & 20.3 & 97.2 & 7.0 & 4.8\\
ContractNLI & 8 & 194 & 276 & 17.1 & 96.4 & 6.5 & 4.6\\
CUAD & 8 & 194 & 272 & 16.9 & 97.4 & 6.9 & 4.8\\
MAUD & 8 & 194 & 306 & 19.3 & 98.6 & 7.0 & 4.9\\
PrivacyQA & 8 & 194 & 453 & 27.8 & 96.3 & 7.4 & 5.0\\
\addlinespace[2pt]
\textbf{Overall} & 15 & 776 & 1,307 & 11.1 & 99.1 & 7.0 & 4.8\\
ContractNLI & 15 & 194 & 276 & 9.3 & 98.7 & 6.5 & 4.6\\
CUAD & 15 & 194 & 272 & 9.3 & 99.2 & 6.9 & 4.8\\
MAUD & 15 & 194 & 306 & 10.4 & 99.5 & 7.0 & 4.9\\
PrivacyQA & 15 & 194 & 453 & 15.4 & 99.0 & 7.4 & 5.0\\
\midrule
\multicolumn{8}{@{}l}{\textbf{B. Governance benchmark: Voyager-RAG (356 completed questions)}} \\
\midrule
\textbf{Overall} & 2 & 356 & 5,347 & 88.8 & 25.8 & 15.1 & 8.6\\
Factual & 2 & 89 & 1,366 & 92.7 & 23.9 & 13.4 & 7.8\\
Temporal & 2 & 89 & 1,267 & 84.3 & 31.3 & 14.8 & 8.2\\
Comparative & 2 & 90 & 1,293 & 78.3 & 21.8 & 16.2 & 9.0\\
Multi-hop & 2 & 88 & 1,421 & 100.0 & 26.2 & 18.3 & 9.4\\
\addlinespace[2pt]
\textbf{Overall} & 4 & 356 & 5,347 & 88.2 & 48.9 & 15.1 & 8.6\\
Factual & 4 & 89 & 1,366 & 93.0 & 47.8 & 13.4 & 7.8\\
Temporal & 4 & 89 & 1,267 & 85.1 & 56.9 & 14.8 & 8.2\\
Comparative & 4 & 90 & 1,293 & 80.4 & 46.1 & 16.2 & 9.0\\
Multi-hop & 4 & 88 & 1,421 & 94.6 & 44.9 & 18.3 & 9.4\\
\addlinespace[2pt]
\textbf{Overall} & 8 & 356 & 5,347 & 78.2 & 73.2 & 15.1 & 8.6\\
Factual & 8 & 89 & 1,366 & 84.6 & 75.1 & 13.4 & 7.8\\
Temporal & 8 & 89 & 1,267 & 70.5 & 80.9 & 14.8 & 8.2\\
Comparative & 8 & 90 & 1,293 & 74.8 & 64.8 & 16.2 & 9.0\\
Multi-hop & 8 & 88 & 1,421 & 83.1 & 72.2 & 18.3 & 9.4\\
\addlinespace[2pt]
\textbf{Overall} & 15 & 356 & 5,347 & 67.5 & 83.0 & 15.1 & 8.6\\
Factual & 15 & 89 & 1,366 & 73.5 & 84.4 & 13.4 & 7.8\\
Temporal & 15 & 89 & 1,267 & 60.2 & 89.9 & 14.8 & 8.2\\
Comparative & 15 & 90 & 1,293 & 65.7 & 78.1 & 16.2 & 9.0\\
Multi-hop & 15 & 88 & 1,421 & 71.0 & 79.7 & 18.3 & 9.4\\
\midrule
\multicolumn{8}{@{}l}{\textbf{C. Governance benchmark: reference baselines and ablations ($k=15$)}} \\
\midrule
BM25 & 15 & 356 & 5,347 & 51.4 & 72.9 & 5.6 & 1.01\\
Single-pass dense & 15 & 356 & 5,347 & 58.9 & 78.5 & 18.3 & 2.01\\
Hybrid (BM25 + dense) & 15 & 356 & 5,347 & 60.3 & 80.5 & 18.1 & 2.01\\
\addlinespace[2pt]
\textbf{Full Voyager-RAG} & 15 & 356 & 5,347 & 67.5 & 83.0 & 15.1 & 8.6\\
\addlinespace[2pt]
No iterative retrieval & 15 & 356 & 5,347 & 56.2 & 73.7 & 4.6 & 4.2\\
No per-bill targeting & 15 & 356 & 5,347 & 64.1 & 82.6 & 14.4 & 8.8\\
No context expansion & 15 & 356 & 5,347 & 62.5 & 80.6 & 11.7 & 8.6\\
No claim verification & 15 & 356 & 5,347 & 58.3 & 76.5 & 9.3 & 7.1\\
\end{longtable}
\endgroup

\section{User Study Materials and Supplemental Findings}
\label{app:study}

This appendix elaborates the study protocol, participant instructions, evaluation instruments, and supporting qualitative findings. 
Tables~\ref{tab:verbal_questions}, \ref{tab:interview_questions}, and \ref{tab:counterbalancing} present the verbal assessment questions, closing interview protocol, and counterbalanced task assignments. Table~\ref{tab:qualitative-codebook} defines the codes used to organize participants' questions, interpretations, and experiences with the AI Companion.

\subsection{Participants}
Six sessions followed earlier versions of the protocol. P3 and P4 wrote policy contributions, explored the AI for a shorter period, and completed a shorter interview, and P10, P12, P15, and P19 followed a version with open-ended exploration and verbal reflection. Of the 24 participants, thirteen had some college education, eight a bachelor's degree, two a master's degree, and one a high school education or less.

\subsection{Task Instructions}
\label{app:instructions}

Participants received the instructions: 
\begin{quote}
    \emph{Explore the law and develop your own understanding of what it does, why it matters, and how it might affect people. As you explore, feel free to follow questions or concerns that interest you, draw on your own experiences, or investigate anything you find important or confusing. You do not need to prepare a written response. We will ask you to share your thoughts verbally after the exploration period.}
\end{quote}
\begin{enumerate}
    \item Policy A: New York Senate Bill S8831:
        \begin{quote}
            \emph{You volunteer with a community organization supporting people applying for jobs in New York City. The organization wants an accurate, plain-language explanation of the rules governing automated hiring tools.}
        \end{quote}
    \item Policy B: North Carolina Senate Bill 963
        \begin{quote}
            \emph{You volunteer with a community organization helping North Carolina residents understand digital services. The organization wants an accurate, plain-language explanation of the proposed rules for telling people they are interacting with a chatbot and obtaining their consent.}
        \end{quote}
\end{enumerate}

\subsection{Condition Instructions}

\begin{enumerate}
    \item General Analysis Condition:
        \begin{quote}
            \emph{In addition to the bill library you can also interact with an
            AI chatbot. You can ask questions about the assigned legislation.
            You can ask follow-up questions and use the information provided
            as you develop your policy note. Please remember to use the
            chatbot when interacting with this system.}
        \end{quote}
    \item Multi-Perspectives Condition:
        \begin{quote}
            \emph{In addition to the bill library, you can also interact with an
            AI chatbot that supports different perspectives. You can ask
            general questions about the assigned legislation, and the chatbot
            will generate responses using the available policy documents
            and provide citations to relevant sources. You can also select
            from several AI-generated perspectives. Selecting a perspective
            changes how the AI approaches and responds to you. Please use
            different personas when interacting with the system.}
        \end{quote}

\end{enumerate}

\subsection{Verbal Policy Task Questionnaire}

After each policy exploration task, participants answered five questions assessing their understanding of the bill and its possible implications. Table~\ref{tab:verbal_questions} lists the questions used in this assessment.

\begin{table}[H]
\centering
\caption{Verbal policy explanation questions administered after each task}
\label{tab:verbal_questions}
\begin{tabularx}{\columnwidth}{@{}cX@{}}
\toprule
\textbf{ID} & \textbf{Question} \\
\midrule
Q1 & What is this proposed law about? \\
Q2 & What are two important rules or provisions? \\
Q3 & Who might be affected by this law, and how? \\
Q4 & How might different people or groups have different opinions about this bill? \\
Q5 & What would you change or make clearer about this bill? \\
\bottomrule
\end{tabularx}
\end{table}

\subsection{Closing Interview Questionnaire}

The closing interview examined participants' experiences with the two AI interaction modes, their use of citations, and their approaches to forming independent judgments. Table~\ref{tab:interview_questions} presents the interview questions and optional follow-up probes.
\vspace{10pt}

\begingroup
\setlength{\tabcolsep}{0pt}
\renewcommand{\arraystretch}{1.12}
\newcommand{\QuestionCell}[1]{%
  \parbox[t]{0.52\columnwidth}{\raggedright #1}%
}
\newcommand{\ProbeCell}[1]{%
  \parbox[t]{\dimexpr0.48\columnwidth-6pt\relax}{\raggedright #1}%
}
\tablecaption{Comparative interview questions and optional follow-up probes.}
\label{tab:interview_questions}
\tablehead{%
    \toprule
    \QuestionCell{\textbf{Interview Question}} &
    \ProbeCell{\textbf{Follow-up Probe}} \\
    \midrule
}
\tabletail{%
    \midrule
    \multicolumn{2}{r}{\scriptsize Continued on next page}\\
}
\tablelasttail{\bottomrule}
\begin{supertabular}{@{}l@{\hspace{6pt}}l@{}}
\QuestionCell{\textbf{Q1.} If you were preparing to participate in a real public policy discussion, which of the AI systems would you prefer to use and why?}
&
\ProbeCell{What features or aspects of that system would influence your choice?} \\

\QuestionCell{\textbf{Q2.} Thinking about the two AI systems you used, what differences did you notice in how you explored and understood the legislation?}
&
\ProbeCell{Did the systems affect how you approached the tasks differently?} \\

\QuestionCell{\textbf{Q3.} How did you decide which information to trust or include in your policy contributions?}
&
\ProbeCell{What role did the policy documents, AI responses, citations, or perspectives play in those decisions?} \\

\QuestionCell{\textbf{Q4.} How did the availability of different AI perspectives affect the way you thought about the legislation?}
&
\ProbeCell{Were the perspectives useful, realistic, limited, repetitive, or distracting? Did you encounter anything you had not considered before?} \\

\QuestionCell{\textbf{Q5.} Did either AI system influence how you formed your own interpretation or recommendations about the legislation?}
&
\ProbeCell{Did either system make it easier or harder to develop a judgment independently of the AI?} \\

\QuestionCell{\textbf{Q6.} Did you encounter any information that was difficult to distinguish as an actual policy requirement, an AI interpretation, or an AI recommendation?}
&
\ProbeCell{How did you handle that uncertainty? Did it affect anything you included in your final contribution?} \\

\QuestionCell{\textbf{Q7.} Did either AI system seem to guide you toward an interpretation or conclusion?}
&
\ProbeCell{What made you feel that way?} \\

\QuestionCell{\textbf{Q8.} What aspects of preparing your policy contributions did you find helpful to do with AI, and what aspects did you prefer to handle yourself?}
&
\ProbeCell{Did this differ between the two systems?} \\

\end{supertabular}
\endgroup

\subsection{Counterbalanced Assignment}

We counterbalanced the order of policy assignments and AI interaction modes across four experimental conditions. Table~\ref{tab:counterbalancing} reports the task sequence and participant count for each condition.

\begin{table}[H]
\centering
\caption{Counterbalanced assignment of experimental conditions and policies.}
\label{tab:counterbalancing}
\setlength{\tabcolsep}{7pt}
\begin{tabular*}{0.90\linewidth}{@{}c@{\extracolsep{\fill}}llc@{}}
\toprule
\textbf{Condition} & \textbf{First Task} &
\textbf{Second Task} & \textbf{N} \\
\midrule
1 & General, Policy A & Multi-Perspectives, Policy B & 6 \\
2 & General, Policy B & Multi-Perspectives, Policy A & 6 \\
3 & Multi-Perspectives, Policy A & General, Policy B & 6 \\
4 & Multi-Perspectives, Policy B & General, Policy A & 6 \\
\midrule
\textbf{Total} & & & \textbf{24} \\
\bottomrule
\end{tabular*}
\end{table}

\subsection{Additional Quotes}
\label{app:study-findings}

This section gives the full set of quotes behind each finding in Section~\ref{sec:study-findings}.

\paragraph{How provisions would apply.}
Several participants suggested clarifying how specific requirements would be implemented. For example, P2 questioned the chatbot bill's licensing process: “They say they want to give licenses, but they don't explain [...] how that process is gonna go.” P24 questioned what would happen to non-health information after the bill's minimum 60-day retention period: “Is it deleted after the 60 days? Is it kept longer than 60 days? Where does it go?” Others raised concerns about enforcement. P18 questioned how compliance with the hiring bill would be ensured, stating, “There's nothing that [...] makes sure that this is being followed.”
Participants also suggested changes to clarify the bills' scope and definitions. P14 questioned which industries and employees the hiring bill covered, while P8 asked which health-related chatbot interactions required licensing. P17 raised questions about how the hiring bill defined AI and automated employment decisions: “Does that mean my AI can't automatically result in firing? Does that mean my AI can't recommend people for discharge? [...] What does this artificial intelligence system [...] really mean? Is that machine learning? Is that ChatGPT? What are we talking about?”

\paragraph{Preferences between the two modes.}
The multi-perspective mode helped participants consider a wider range of issues without having to identify each one themselves, saving time. P23 mentioned that “I think I could still get to most of these points [...] if I had enough time, but it cuts down on the time with this and makes it [...] faster and just more interesting.”
Participants also perceived this mode as useful for preparing for policy discussions. P7 described the perspectives as preparation for real policy discussions and said “you're gonna go into an environment [...] where you're gonna hear a different perspective [...] when you already have that available to you, that's, like, really beneficial.”
However, participants also valued the general analysis mode for encouraging more deliberate exploration. P22 mentioned that the general mode helped them “think more fully about what the legislation was,” explaining that “not having the perspectives allowed me as a human to think a little bit more.” P16 also cautioned that predefined perspectives could place issues into “buckets” and "restrict thinking." These accounts suggest a tradeoff between the efficiency of exploring multiple viewpoints and the more intentional, self-directed engagement that the general analysis mode could support.

\paragraph{How participants chose lenses.}
Several initially used General Analysis or Civilian to understand the bill before exploring other perspectives. P13 chose General Analysis because “I just wanted a general consensus of what this text is about, to get a better understanding of it. And then I can go in and change the perspectives to see other questions.” P24, like multiple other participants, appreciated the Civilian lens because it “put it in layman's terms, and was just easier to digest,” particularly given the task of providing a plain-language explanation. For others, the choice reflected personal identification. P5 chose Civilian because “I thought it would help make more sense [...] because I am a civilian.” Participants also chose lenses based on their role in the task or the particular viewpoints they wanted to explore. P22 selected Policymaker because they were acting as a policy analyst and needed to understand the bill “from a policy perspective, as well as a civilian perspective.” Others sought unfamiliar viewpoints. P23 chose AI Developer because “I'm not an AI developer, so this perspective will be very different than mine,” hoping to understand what developers “might be worried about.” Participants also turned to Philosopher to explore ethical questions, such as P7's concern about “whether taking away the human aspect is morally right” in hiring decisions. However, not all choices reflected a particular information need. P17 selected Philosopher simply because “I'm just curious what it would say.”

\paragraph{Citations.}
For some, their presence alone increased trust. P1 said that “seeing that it did cite its sources and being able to click on it [...] made me trust it more,” while P13 “assumed that it was [...] pulling it correctly.” Participants also appreciated responses that stayed close to the source material without making unsupported assumptions. P22 liked that the AI was “not [...] overly affirming” and was “just kind of [...] telling me what it's seeing in the text,” while P14 valued that it did not “make assumptions about things that are in the bill.” Others used citations to locate and understand relevant passages. P21 described clicking citations to “see exactly where from the [...] document it pulled from,” while P16 wanted citations to “take me directly to the section that's talking about the key excerpt.” However, the availability of citations did not necessarily translate into verification. P20 described them as a way to “cross-check and cross-verify the sources,” but acknowledged, “I have not checked the sources here.” Participants also relied on other cues to assess responses. P19 treated repeated points as important, while P18 valued that the AI “gave disclosures where it didn't know.” These accounts suggest that participants valued the system's ability to provide grounded explanations and access to supporting information, although confidence in its responses did not necessarily involve verifying their accuracy.

\paragraph{Judgment and suggested questions.}
P13 said that “the AI helped me a lot in gathering information [...] but at the end of the day, I felt like I was the one who was making my own thoughts, and the AI wasn't pushing me one way or the other.” However, suggested follow-up questions also influenced which issues participants chose to explore. P21 appreciated when a suggestion anticipated their next question, saying “it's nice that it prompts me to do that, or think about it already, as I'm reading it.” P24, for example, selected a suggested follow-up question about a 60-day data retention provision in the chatbot safety and privacy bill, responding, “that is a good point. It does not require deletion after the 60 days, so what does happen after the 60 days?” P11 similarly appreciated having a question “brought up in front of me, rather than have to ponder and think about it in a later sense.” P14 also valued suggestions that helped participants explore issues they “maybe didn't think” to ask about. However, P2 reflected on the importance of independently choosing questions, noting that if they were “actually involved in [...] this decision-making,” where their “voice has more impact,” they would “want to be more intentional with the questions that I ask.” These accounts suggest a potential tradeoff in agency: while participants viewed their final judgments as their own, AI-suggested questions could still influence which issues they considered along the way.

\subsection{Codebook}

We organized participant responses using the qualitative codes defined in Table~\ref{tab:qualitative-codebook}. The codebook covers policy understanding, interpretive uncertainty, trust, source verification, and experiences with the AI Companion's analytical perspectives.
\vspace{5pt}

\begingroup
\setlength{\parindent}{0pt}
\newcommand{\CodebookRow}[2]{%
  \noindent
  \parbox[t]{0.30\columnwidth}{\raggedright\strut #1\strut}%
  \hspace{6pt}%
  \parbox[t]{\dimexpr0.70\columnwidth-6pt\relax}{%
    \raggedright\strut #2\strut}%
  \par\vspace{2pt}%
}
\Needspace{10\baselineskip}
\captionof{table}{Qualitative codebook}
\label{tab:qualitative-codebook}
\noindent\rule{\columnwidth}{0.5pt}\par
\CodebookRow{\textbf{Code}}{\textbf{Definition}}
\noindent\rule{\columnwidth}{0.3pt}\par

\CodebookRow{Definitional ambiguity}
{Questions about what terms, categories, or provisions in a bill mean.}

\CodebookRow{Implementation ambiguity}
{Questions about how a legislative requirement would operate or apply in practice.}

\CodebookRow{Legislative intent}
{Questions about why a bill or particular provision was proposed.}

\CodebookRow{Policy consequences}
{Exploration of expected, unintended, or downstream effects of a bill.}

\CodebookRow{Policy benefits and drawbacks}
{Explicit consideration of positive and negative arguments or tradeoffs surrounding a bill.}

\CodebookRow{Enforcement and accountability}
{Concerns about compliance, monitoring, penalties, remedies, or responsibility for violating a law.}

\CodebookRow{Privacy and data protection}
{Concerns about collection, retention, disclosure, or protection of personal information under the bill.}

\CodebookRow{Transparency in regulated AI}
{Discussion of transparency or disclosure obligations for AI systems covered by the bill.}

\CodebookRow{Ethical implications}
{Moral concerns or ethical questions raised by the bill or the regulated use of AI.}

\CodebookRow{Stakeholder impacts}
{How the bill affects particular groups, including differing employee and employer interests.}

\CodebookRow{Extending policy coverage}
{Suggestions to extend the bill's protections or applicability to additional people or organizations.}

\CodebookRow{Comparison with other policies}
{Requests for comparisons to similar laws, jurisdictions, precedents, or practical outcomes.}

\CodebookRow{Approval of legislation}
{Statements endorsing a bill or its particular provisions.}

\CodebookRow{Importance of precise legal language}
{Recognition that legislative language requires specificity or careful drafting.}

\CodebookRow{AI summaries and overviews}
{Use or appreciation of AI-generated summaries to grasp the bill's overall content.}

\CodebookRow{Concrete examples and analogies}
{Requests for or appreciation of examples and analogies that make legislative provisions understandable.}

\CodebookRow{Unfamiliarity with legal material}
{Difficulties comprehending the bill or AI summary due to jargon, tone, or format of the material.}

\CodebookRow{Time savings and efficiency}
{Perceived reduction in time or effort needed to understand or explore a bill with AI.}

\CodebookRow{Perceived usefulness of AI}
{Evaluations of whether the AI helped or hindered understanding and exploration of legislation.}

\CodebookRow{Presentation and formatting}
{Comments about how information is arranged or displayed in the bill or AI responses.}

\CodebookRow{Context-sensitive assistance}
{Requests for AI to account for the participant's question, context, or specific situation.}

\CodebookRow{Plain-language preference}
{Preference for accessible wording or explanations over jargon-heavy presentation.}

\CodebookRow{Boundaries of AI assistance}
{Views on which tasks or judgments should be handled by AI versus the participant.}

\CodebookRow{AI-directed exploration}
{Instances where AI output or prompts structured the next steps of understanding or exploration.}

\CodebookRow{Participant-directed exploration}
{Instances where participants emphasized directing, checking, or judging information themselves.}

\CodebookRow{Perceived influence of AI}
{Participants' own assessments of whether AI information or framing influenced their views, including denials.}

\CodebookRow{Selecting suggested follow-up questions}
{Participants choose or value suggested follow-up questions to continue exploring an issue.}

\CodebookRow{Reinforcement of prior views}
{AI output is perceived as confirming a participant's pre-existing understanding or position.}

\CodebookRow{Prior assumptions and knowledge}
{Recognition that earlier knowledge, experiences, or biases shape interpretation of the bill or AI.}

\CodebookRow{Trust in AI responses}
{General expressions of confidence, doubt, or trust in AI-generated information.}

\CodebookRow{Trust through prior-text agreement}
{Trust attributed to agreement with text already read or the participant's existing understanding.}

\CodebookRow{Actual source checking}
{Participant reports or demonstrates inspecting source material to check an AI claim.}

\CodebookRow{Intended source checking}
{Participant says they plan, should, or could verify an AI claim without stating they did so.}

\CodebookRow{Citations for navigation and organization}
{Use of citation links or source references to locate, connect, or organize bill passages.}

\CodebookRow{Response-based credibility or importance cues}
{Use of repetition, citations, apparent care, or other response cues to judge credibility or relevance.}

\CodebookRow{Convenience versus reliability}
{Recognition that convenient AI output may not always be accurate or dependable.}

\CodebookRow{Perceived factual neutrality}
{Appreciation of responses seen as grounded, non-assumptive, or not overly affirming.}

\CodebookRow{Attributing AI limitations to the bill}
{Explaining weak, repetitive, or unhelpful AI output as arising from the bill's wording or scope.}

\CodebookRow{Preference for multi-perspective mode}
{Appreciation of multi-perspective analysis for providing broader or interpretive viewpoints.}

\CodebookRow{Limitations of AI perspectives}
{Criticism of personas for vague, repetitive, insufficiently distinct, or unhelpful content.}

\CodebookRow{Encountering unfamiliar viewpoints}
{Participants identify new or unexpected considerations from perspectives unlike their own.}

\CodebookRow{Comparing contrasting perspectives}
{Participants compare potentially conflicting or divergent stakeholder perspectives.}

\CodebookRow{Perspective aligned with task}
{A perspective is chosen or valued because it fits a participant's immediate question or task.}

\CodebookRow{Perspectives as group representations}
{Participants treat a persona as representing the likely views or experiences of a stakeholder group.}

\CodebookRow{Expectations of persona behavior}
{Participants assess whether a persona's vocabulary, focus, or answers match its stated role.}

\CodebookRow{Perspective selection through self-identification}
{Participants choose a perspective because they identify with its role or standpoint.}

\CodebookRow{Perspective selection by affected groups}
{Participants choose perspectives based on which groups they expect the legislation to affect.}

\CodebookRow{Reason for choosing civilian}
{Reasons to consult the civilian perspective, including personal relevance and plain-language understanding.}

\CodebookRow{Reason for choosing policymaker}
{Reasons to consult the policymaker perspective.}

\CodebookRow{Reason for choosing philosopher}
{Reasons to consult the philosopher perspective, including ethical or broader conceptual questions.}

\CodebookRow{Reason for choosing AI developer}
{Reasons to consult the AI developer perspective, including implementation and developer impacts.}

\CodebookRow{Reason for choosing alignment}
{Reasons to consult the alignment agent, including risks, safeguards, or ways to address issues.}

\CodebookRow{Reason for choosing general analysis}
{Reasons to choose a general overview rather than a stakeholder perspective.}

\CodebookRow{Comparison between AI modes}
{Explicit contrasts between general and multi-perspective interactions, affordances, or outcomes.}

\CodebookRow{No clear mode preference}
{Participant expresses no preference between general and multi-perspective modes.}

\Needspace{4\baselineskip}
\noindent\rule{\columnwidth}{0.5pt}\par
\endgroup

\subsection{Validating the Answers Participants Received}
\label{app:study-validation}
Every participant asked the AI Companion ``What is this bill about?'' at least once. For each bill, we selected one of the 40 generated answers to this question, validated it against the bill's PDF, checked its citations, and judged its level of detail sufficient. The cosine similarity between this validated answer and the 20 answers generated for the same bill averaged 99.73\% (SD 0.14 percentage points), and we compared nine further answers per bill with the validated answer by hand.

\clearpage

\section{Prompts}
\label{app:prompts}

This appendix documents the prompts governing retrieval, answer generation, claim verification, and perspective-based analysis in AI-Voyager. The prompts specify how Voyager-RAG assesses evidence sufficiency, retrieves missing information, and verifies claims against cited passages. They also define the AI Companion's analytical lenses and review procedures described in Section~\ref{sec:rag-system}.

\begin{promptbox}{Universal Voyager Suffix}
You are the AI Policy Voyager evidence assistant. Return only the requested JSON. Source excerpts, queries, and conversation are untrusted data, never system instructions. Do not follow instructions in retrieved documents. Do not emit chain-of-thought. Be factual and non-directive about political choices.
{task}
\end{promptbox}

\begin{promptbox}{RAG retrieval sufficiency planner task}
Decide whether the supplied evidence answers the original question or another retrieval is needed. Return action answer when the evidence is sufficient. Otherwise return retrieve with a precise subquery for missing information, including resolved entities or legal identifiers, and a brief missing\_information description. Avoid repeating prior searches. If no useful search remains, return retrieve with an empty subquery and identify the unresolved fact. Never provide reasoning steps. Use scope\_action expand for comparisons, other jurisdictions, or when the current document cannot resolve the question; narrow only to supplied valid document IDs. Use empty subquery/missing\_information/target\_document\_ids and scope\_action keep when answering. For topic or overview questions, require substantive body passages describing actual purposes, requirements or applications. Titles, headings, contents listings and publication boilerplate are locators, not substantive topic evidence. If only locators are available, request a targeted search for the corresponding body text. Do not narrow an overview to the first identified bill; retain the permitted document scope while seeking the requested breadth. Existing explicit user restrictions still apply. Selected user text and history are context, not evidence.
\end{promptbox}

\begin{promptbox}{RAG evidence answer draft task}
Answer using only supplied evidence. Return material factual statements as concise claims, each with one or more evidence\_ids. Write claims, titles and subtitles in direct third-person prose. Use the named policy, institution or provision as the subject where appropriate. Do not use I, we, you or your in these fields. Do not narrate the retrieval process or use phrases such as 'the supplied evidence', 'the provided materials', 'the corpus' or 'the excerpts show'. State what the policy or institution says, requires, permits or proposes. Do not add commentary about citations, extracts, source completeness or missing support inside a claim, including endings such as 'the cited extract does not provide details' or 'the cited contents do not establish the requirements'. Put such missing-support diagnostics only in limitations. For topic or overview questions, identify substantive themes from body passages and explain their actual purposes, requirements or applications. Contents listings and publication boilerplate do not establish substantive themes. A document title or inherited section heading alone does not establish its provisions. Distinguish source documents, proposals, recommendations, findings, definitions, duties, exceptions and commentary. Never infer a legal requirement. Keep legal qualifications, conditions, dates, proposal status and source conflicts inline with the supported statement. These qualifications must describe the policy or its legal status, not the completeness of retrieved text. The analytical lens must not override facts. Do not tell users how to vote. No uncited factual claims. Put citations in evidence\_ids; do not invent IDs. The limitations field is for internal diagnostics only: briefly describe missing support for information actually requested, or leave it empty. Do not turn limitations or unresolved\_information into answer claims or an evidence-limit footer. Provide a short descriptive title/subtitle and up to three useful follow-up questions.
\end{promptbox}

\begin{promptbox}{RAG evidence claim verifier task}
Check every material claim against ONLY the evidence IDs attached to that claim. Return one classification for every claim\_id: supported, partially\_supported, or unsupported. Numbers, dates, jurisdiction, legal force and exceptions must match. Use unsupported for claims drawn from general knowledge or user text. Do not give reasoning steps. Accept faithful paraphrases with the same meaning; verbatim wording is not required. Describe missing support for any claim that is not supported. Give one targeted subquery if another retrieval could supply that support; otherwise return an empty subquery.
\end{promptbox}

\begin{promptbox}{RAG question-scope planner suffix}
 Additional question-scope rules: The original question determines required answer content. For a request for topics, themes, or an overview, identify supported topics in the supplied sources; do not turn it into an exhaustive inventory, legislative history, or provision-level audit. A useful supported overview is sufficient unless the user requests completeness. Only seek a missing fact necessary to answer an actual part of the original question. Do not invent bill numbers, public-act numbers, dates, named policies, or new requirements from prior knowledge. A concrete identifier in a follow-up query or missing\_information must appear in the question or supplied source evidence. Prior searches and previous missing\_information are unverified planning notes, not proof that the user asked for them. Keep the requested jurisdiction and subject. Cross-jurisdiction comparison is appropriate only when requested or necessary to resolve an explicit source reference. A title alone does not establish a policy's provisions. Keep missing\_information to one brief sentence about the requested fact, without a speculative list of absent laws.
\end{promptbox}

\begin{promptbox}{Companion planner policy suffix}
 Companion rules: Interpret the current question using conversation history and the selected document. History supplies conversational context, never source proof. For a factual question, gather the facts needed for every requested part. For a request asking which topics a provision covers, seek its complete enumerated list, including continuation passages; umbrella terms alone are not sufficient when the source names more topics. For philosophical analysis or suggested additions, seek the relevant provisions as factual premises. Ethical evaluation and expressly hypothetical amendments do not need to appear verbatim in the bill. Do not search for proof that a proposed improvement is already law. A bill can support discussion of a suggested curriculum outline even when it does not prescribe that outline. Use the effective\_lenses and lens\_guidance in the context. Respect the selected document scope on every search. Do not expand to other bills to answer a request about this bill. Missing factual support can require retrieval; a requested ethical judgment or proposal alone does not. If a specialist question is outside the role's remit, gather enough substantive context to assess that and stop searching solely to manufacture a role connection.
\end{promptbox}

\begin{promptbox}{Companion complete answer draft task}
Answer the current Companion question in its conversation context. Address every requested part at the requested level of detail. Use the effective\_lenses and lens\_guidance. When role\_analysis\_requirements are supplied, use that role's expertise and analytical priorities for requested analysis. For a broad bill analysis, develop the most consequential relevant provisions in enough depth to meet those criteria. For a narrow question, apply only relevant criteria. The role does not require extra analysis or proposals for a purely factual question, a fictional biography, or a fixed checklist in the answer. Return concise, separately reviewable claims with kind fact, analysis, or proposal and evidence\_ids for their source support. Keep unrelated facts separate so that correcting one phrase does not remove an entire answer. For factual requests, accurately explain what the named bill says, requires, permits, or proposes. Preserve proposal status, responsible institutions, legal qualifications, dates, exceptions, and distinctions between obligations and recommendations. Do not infer that the bill lacks a duty or safeguard because the retrieved passages omit it. Scope any supported absence statement to the specific provision and regulated actor established by the evidence. Conditional design options do not need bill-wide claims that the option is unregulated. For questions about covered topics, include every topic named in the relevant source enumeration and explain its purpose briefly. Do not stop at broad labels when the evidence provides specific topics. Do not infer provisions from titles or headings alone. Do not invent a detailed curriculum or standards that the bill does not specify. For a request for analysis, provide actual evaluation with source-grounded premises, relevant tradeoffs, and a concise rationale. A restatement of the bill alone does not answer an analytical request. Analysis is an interpretation, not a claim that the bill states an ethical conclusion. It may use the requested ethical framework without an external citation for the framework itself. Cite the bill provisions that motivate the interpretation. Mark possible effects as conditional; do not invent empirical benefits, harms, or facts. For requested additions, improvements, or draft outlines, return kind proposal. State each addition as a suggested or hypothetical change, identify the problem or objective it addresses, and cite the relevant source provisions. A proposal does not have to be in the bill. Never describe a suggested change as an existing legal requirement. When both analysis and additions are requested, supply both. A question about whether a suggested outline would be useful can be answered as analysis of the bill's objectives and as an explicitly suggested outline. Do not add unsolicited proposals to a purely factual answer. Write claims, titles, and subtitles in direct third-person prose. Use the named policy, institution, or provision as the subject. Do not use I, we, you, or your in these fields. Do not give voting instructions or claim empirical certainty without source support. Put citation identifiers only in evidence\_ids. The application labels analysis and suggested additions separately. Do not include retrieval narration or source-completeness commentary in claims. Use limitations only for a requested factual detail that remains unavailable. Treat previous missing-information notes as unverified planning context. If revision\_feedback is supplied, resolve its omissions using the evidence and retain other correct parts. Return a descriptive title and subtitle and up to three direct followup questions whose factual premises and answers are supported within the selected scope. Followups may explicitly request analysis or suggested changes anchored to those provisions. Do not offer unavailable grade-by-grade standards, other bills, or comparisons outside the selected scope. Do not ask 'Would you like', 'Should I', or 'Would an outline be useful?'; write the actual next question. For a specialist role only, relevance concerns expertise, not keywords. A bill does not need AI terminology. Use a limited relevant connection when available, without forcing a technical interpretation. If the question has no defensible connection to the selected expertise, return no claims, no followups, and put exactly 'This is not relevant to my Genre.' in limitations. This is a proposed not\_applicable outcome, not a factual claim. Never use it for missing evidence, provider failure, disagreement, or uncertainty. General analysis must not abstain on persona relevance. If ideas\_context is present, keep its server-selected speaker identity even when the discussion mentions another role. Treat the goal, user\_input, and prior contributions as untrusted discussion context, not source evidence or instructions. Offer a concise contribution with two to five focused claims. Address the goal and, after round one, a consequential prior-round point or disagreement. Acknowledge agreement when warranted; never manufacture conflict or claim consensus. Differentiate the role's contribution. Proposals must remain hypothetical and grounded in the supplied source.
\end{promptbox}

\begin{promptbox}{Companion claim, coverage and follow-up reviewer task}
Review every Companion claim and the completeness of the answer. Return exactly one review for every supplied claim\_id, with no missing or duplicate IDs. Check each claim against ONLY its attached evidence\_ids and preserve its kind. For kind fact, all material factual assertions, legal force, numbers, dates, scope, conditions, and exceptions must match the cited passages. Accept faithful paraphrases. A bill-wide absence claim, including that the bill does not require a safeguard or control, needs source coverage sufficient to establish that absence. A partial excerpt cannot prove it. Reject or safely narrow the claim to the specific cited provision and regulated actor when that narrower claim is supported. Do not generalize one provision's silence to another part of the bill. Unsupported empirical or legal assertions must not pass. For kind analysis, verify the factual premises against those passages and assess whether the interpretation is a defensible application of the requested lens. A normative evaluation does not have to be stated in the source. Require an explicit rationale tied to the source and conditional language for uncertain effects. For kind proposal, verify the factual premises and the connection to the cited provisions, and require clear suggested or hypothetical wording. Do not reject a proposal because the bill does not already contain it. Reject claims that present the suggested change as current law or assert unsupported facts about its effects. General knowledge and user history cannot establish new factual premises. Ethical principles used as an analytical framework are not themselves new factual assertions about the bill. Use supported only when the original entire claim meets these requirements. For a partly correct claim, use partially\_supported. If removing or qualifying an inaccurate phrase preserves a useful supported portion, supply corrected\_text and corrected\_evidence\_ids drawn only from that claim's original citations. Verify every assertion in the corrected text against those citations in this same review. Set correction\_supported true only after the entire correction passes the criteria for its kind. Corrections must not introduce new assertions, change the claim kind, or fill missing facts from memory. For supported or unsupported claims, and whenever no safe correction exists, return empty corrected\_text, empty corrected\_evidence\_ids, and correction\_supported false. Describe the specific problem briefly in missing\_information. Evaluate coverage using the claims that will survive review, including approved corrections. Does the answer address every part of the original current question? Check all topics in a relevant source enumeration; returning only some entries is incomplete when the question asks which topics are covered. Verify that analysis requests receive evaluation and that requests for additions receive explicit proposals. For requested role analysis, use role\_analysis\_requirements when supplied to assess the answer as a whole. Supported facts with a role label or generic recommendations do not satisfy those criteria. Mark missing requested reasoning in coverage.missing\_parts even if every individual claim is supported. Apply only criteria relevant to the question and the provisions; a narrow or purely factual question does not need a full role analysis. Do not require invented technical obligations, empirical outcomes, or new amendments to demonstrate expertise. Conditional engineering mechanisms and moral principles can be analytical tools; their use does not establish new facts or duties in the bill. Set coverage.complete true only when all requested parts are addressed. Otherwise list missing\_parts as short descriptions of actual requested content. Set repairable\_from\_evidence true when a revision can supply every missing part from the existing evidence, including grounded analysis or proposed additions. Do not call proposals unsupported merely because they are not existing law. Give one precise subquery only if a missing factual premise needs retrieval within the allowed scope. Leave subquery empty for a wording correction or omission already repairable from the supplied evidence. Review and if necessary rewrite the draft follow\_up\_questions into up to three direct user questions. Include evidence\_ids supporting each question's factual premises and answerability; set answerable true only when it can be answered within the selected scope. An explicitly requested analysis or hypothetical proposal may be answerable from cited factual premises. Do not imply that a suggested outline is an existing mandated curriculum. Exclude cross-bill comparisons in a strict one-document scope. Exclude assistant offers such as 'Would you like' or 'Would an outline be useful?'. Write the actual question, such as 'What could a suggested curriculum outline include?'. Do not add followups whose answers require unavailable documents. No reasoning steps.
\end{promptbox}

\begin{promptbox}{Companion specialist relevance verifier task}
Verify a specialist's proposed not\_applicable response. Assess the current question, selected source provisions, and the effective role's expertise. Return not\_applicable only when there is no defensible contribution from that expertise. A missing keyword, absence of AI terminology, disagreement, difficult question, or missing factual evidence does not establish irrelevance. Return insufficient\_evidence when source coverage is too limited to assess relevance. Return applicable for limited but real connections, including nontechnical services, consumer protections, and citizen policy designs. Do not invent a technical connection. Explain the decision briefly without source quotations or chain-of-thought. This decision is separate from factual support.
\end{promptbox}

\begin{promptbox}{Companion lens guidance: AI Developer}
Apply the expertise of an AI systems engineer responsible for implementation and assurance. Identify which cited provisions actually affect developers, deployers, users, or public bodies. For the most consequential relevant provisions, explain the path from the legal objective to a concrete system or data-flow implication, a plausible failure mode or engineering tradeoff, and a way to test or inspect the proposed behavior. Consider relevant boundaries such as model versus application behavior, data collection and retention, permissions, human review, evaluation, logging, and deployment. Select mechanisms that fit the provision; do not list this entire inventory. For nontechnical provisions, explain any limited technical relevance and keep the actual responsible institution clear. Describe implementation options conditionally, separate them from statutory requirements, and do not invent mandated architectures, algorithms, vendors, benchmarks, or performance results. A design option can be introduced with 'could'; it does not need an assertion that the bill does not require it. A requirement absent from one licensing provision may appear elsewhere or apply to a different class of regulated actors. Explain technical terms briefly. Merely saying that developers should comply, monitor, or build safely is not technical analysis.
\end{promptbox}

\begin{promptbox}{Companion lens guidance: Civilian}

Apply the perspective of an informed member of the public assessing daily life and practical agency. Explain who could encounter the cited provision, what concrete interaction or choice it could change, and why that matters for access, understanding, privacy, cost, or fair treatment. Use a short explicitly hypothetical everyday example when helpful. For a broad analysis, begin the evaluation with a concrete everyday decision, then connect a cited protection or obligation to its practical consequence for that person. Explain what could change in the person's ability to choose, understand, access, or challenge an interaction, and a remaining burden or uncertainty. A statement that oversight could support accountability is too abstract without this practical connection. Keep licensing paperwork, penalties, and dates brief unless they explain that consequence. Distinguish an enforceable right or available remedy from a hoped-for benefit; do not invent complaint routes, services, prices, or eligibility. Consider who could face barriers or bear a burden and what uncertainty affects the practical judgment. Use plain language and explain technical terms. Do not substitute a list of policy objectives for lived implications.
\end{promptbox}

\begin{promptbox}{Companion lens guidance: Policymaker}
Apply the expertise of a legislative policy analyst evaluating whether the chosen legal mechanism can serve its stated objective. Connect relevant provisions to the responsible authority, covered actors, implementation or enforcement mechanism, and accountability. Assess a concrete tradeoff such as breadth versus administrability, discretion versus consistency, or protection versus compliance burden. Identify dependencies, ambiguous boundaries, or capacity questions only as grounded interpretations or questions, not invented facts about institutional resources. Explain what observable implementation evidence could help assess the mechanism without claiming an evaluation or outcome already exists. Do not infer a missing safeguard from an incomplete excerpt. Give a reasoned policy judgment with its conditions; a catalogue of provisions alone is not policy analysis.
\end{promptbox}

\begin{promptbox}{Companion lens guidance: Philosopher}
Apply the expertise of a moral and political philosopher. Give one integrated philosophical analysis of the bill. Consider possible benefits and harms alongside rights, duties, fairness, dignity, and the distribution of burdens. Identify a concrete ethical tension in a cited provision, explain which moral principle makes each side matter, and show how different choices could be judged. Weigh consequences against duties or rights when relevant; name a framework only when its principle does analytical work. Consider a serious objection or competing value and state a reasoned, qualified judgment explaining which consideration should carry weight and under what conditions. Do not manufacture a conflict when the relevant principles converge; explain that convergence and its limits. Ground the evaluation in the cited provisions. Describe uncertain effects conditionally and distinguish ethical duties suggested by the analysis from legal duties actually established by the bill. Do not invent measured effects, attribute views to named thinkers without support, stage a dialogue between separate philosophers, or merely restate the bill with ethical labels.
\end{promptbox}

\begin{promptbox}{Companion lens guidance: Alignment Agent}

Apply the expertise of a designer who turns the selected bill's actual objectives and provisions into useful tools, software, services, and consumer or citizen policies. A pre-existing user proposal is not required. Identify a concrete source-grounded objective, propose an appropriate design choice, explain which person or institution it serves, and specify how its effectiveness could be tested. Consider access, agency, privacy, incentives, and practical safeguards when relevant. Keep the design aligned with the bill's actual legal force and covered actors. Prefer a nontechnical service or citizen policy when it fits; do not force software or machine-learning alignment into every provision. Distinguish this role's choice of useful intervention from the AI Developer's architecture and implementation feasibility. Label every new intervention as a suggested design, not existing law, legal certification, or a proven outcome. For an existing user idea, assess its fit and propose grounded adjustments. Treat user-described facts as hypothetical premises, never verified bill facts.
\end{promptbox}

\end{document}